\documentclass[runningheads]{eccv2022submission}
\usepackage{graphicx}
\usepackage{tikz}
\usepackage{comment}
\usepackage{amsmath,amssymb} % define this before the line numbering.
\usepackage{color}

\usepackage{bbm}
\usepackage{algorithm}
\usepackage{multirow}
\usepackage[noend]{algpseudocode}
\usepackage{flushend}
\usepackage{xcolor}
\usepackage{pifont}%
\newcommand{\cmark}{\ding{51}}%
\newcommand{\xmark}{\ding{55}}%
\usepackage{subfig}
\usepackage{wrapfig}

\usepackage{colortbl}

\newcommand{\txt}[1]{{\texttt{#1}}}

\usepackage[accsupp]{axessibility}  % Improves PDF readability for those with disabilities.

\usepackage[pagebackref=true,breaklinks=true,colorlinks,bookmarks=false]{hyperref}

\begin{document}
% \linenumbers
\pagestyle{headings}
\mainmatter
\def\ECCVSubNumber{6665}  % Insert your submission number here

\title{OpenLDN: Learning to Discover Novel Classes for Open-World Semi-Supervised Learning} % Replace with your title

\begin{comment}
\titlerunning{ECCV-22 submission ID \ECCVSubNumber} 
\authorrunning{ECCV-22 submission ID \ECCVSubNumber} 
\author{Anonymous ECCV submission}
\institute{Paper ID \ECCVSubNumber}
\end{comment}
%******************

% \begin{comment}
\titlerunning{OpenLDN: Learning to Discover Novel Classes for Open-World SSL}
% \orcidID{0000-0001-5378-1697}
\author{Mamshad Nayeem Rizve\inst{1} \and
Navid Kardan\inst{1} \and
Salman Khan\inst{2} \and
Fahad Shahbaz Khan\inst{2} \and
Mubarak Shah\inst{1}}
\authorrunning{M. N. Rizve et al.}
\institute{Center for Research in Computer Vision, UCF, USA \and
Mohamed bin Zayed University of AI, UAE \\
\email{\{nayeemrizve, kardan\}@knights.ucf.edu, \{salman.khan, fahad.khan\}@mbzuai.ac.ae, shah@crcv.ucf.edu}
}
% \end{comment}
%******************
\maketitle

\begin{abstract}
Semi-supervised learning (SSL) is one of the dominant approaches to address the annotation bottleneck of supervised learning. Recent SSL methods can effectively leverage a large repository of unlabeled data to improve performance while relying on a small set of labeled data. One common assumption in most SSL methods is that the labeled and unlabeled data are from the same data distribution. However, this is hardly the case in many real-world scenarios, which limits their applicability. In this work, instead, we attempt to solve the challenging open-world SSL problem that does not make such an assumption. In the open-world SSL problem, the objective is to recognize samples of known classes, and simultaneously detect and cluster samples belonging to novel classes present in unlabeled data. This work introduces OpenLDN that utilizes a pairwise similarity loss to discover novel classes. Using a bi-level optimization rule this pairwise similarity loss exploits the information available in the labeled set to implicitly cluster novel class samples, while simultaneously recognizing samples from known classes. After discovering novel classes, OpenLDN transforms the open-world SSL problem into a standard SSL problem to achieve additional performance gains using existing SSL methods. Our extensive experiments demonstrate that OpenLDN outperforms the current state-of-the-art methods on multiple popular classification benchmarks while providing a better accuracy/training time trade-off. Code: \url{https://github.com/nayeemrizve/OpenLDN}%{https://github.com/nayeemrizve/OpenLDN} 
\end{abstract}

\section{Introduction}

Deep learning methods have made significant progress on challenging  supervised learning tasks \cite{he2016deep,szegedy2016rethinking,he2017mask,carreira2017quo,chen2018encoder}. However, the supervised learning paradigm assumes access to large amounts of manually labeled data which is time-consuming and expensive to acquire. Several approaches have been proposed to address this problem, including semi-supervised learning \cite{tarvainen2017mean,miyato2018virtual,NIPS2019_8749_MixMatch}, active learning \cite{NIPS2017_8ca8da41,gal2017deep,sener2018active}, self-supervised learning \cite{doersch2015unsupervised,chen2020simple,he2020momentum}, transfer learning \cite{sharif2014cnn,zamir2018taskonomy,kornblith2019better}, and few-shot learning \cite{pmlr-v70-finn17a,NIPS2016_6385,snell2017prototypical,rizve2021exploring}. Among them, semi-supervised learning (SSL) is one of the dominant approaches which reduces the amount of annotation required by taking advantage of a large collection of unlabeled data.

Even though recent SSL methods \cite{NIPS2019_8749_MixMatch,Berthelot2020ReMixMatch:,FixMatch,xie2019unsupervised} have achieved promising results, one of their primary assumptions is that both labeled and unlabeled data come from the same distribution. However, this assumption is difficult to satisfy in many real-world scenarios (open-world problems e.g.\  \cite{bendale2015towards,kardan2017mitigating}). For instance, unlabeled data is commonly mined from the web sources which %might unintentionally 
can include examples from unknown classes. It has been established that training with such examples generally deteriorates the performance of the standard SSL methods \cite{oliver2018realistic,chen2020semi} . To mitigate the negative impact of unlabeled samples from unknown (novel) classes, different solutions have been proposed \cite{guo2020safe,chen2020semi,zhao2020robust}. However, their main motivation is to merely ignore novel class samples to prevent performance degradation on known classes. In contrast, recently ORCA \cite{cao2022openworld} generalizes the SSL problem with novel classes, where the objective is not only to retain the performance on known classes but also to recognize samples of novel classes. This realistic SSL setup is called open-world SSL problem and is the focus of this work.

This work proposes OpenLDN which employs a pairwise similarity loss to discover novel classes. This loss solves a pairwise similarity prediction task that determines whether an image pair belongs to the same class or not. Essentially, this task is akin to unsupervised clustering problem \cite{chang2017deep,wu2019deep}, thereby promoting novel class discovery by identifying coherent clusters. The fundamental challenge for solving pairwise similarity is to determine similarity relationship between a pair of images without accessing their class labels. One common way to overcome this challenge is to estimate the pairwise similarity relationship based on pretrained unsupervised/self-supervised features \cite{Han2020Automatically,cao2022openworld}. However, this process is computationally expensive. To avoid dependency on unsupervised/self-supervised pretraining, instead, we exploit the information available in the labeled examples from known classes for solving the pairwise similarity prediction task, and introduce a pairwise similarity prediction network to generate the similarity scores between a pair of images. To update the parameters of this network, we resort to a bi-level optimization rule \cite{bard2013practical,guo2020safe}, which transfers the information available in the labeled examples of known classes to utilize them in learning unknown classes. In particular, we \emph{implicitly} optimize the parameters of the similarity prediction network based on the cross-entropy loss on labeled examples. This way, we solve the pairwise similarity prediction task without relying on unsupervised/self-supervised pretraining, which makes the overall training more efficient while providing substantial performance gains. 

Learning pairwise similarity relationship based on output probabilities leads to implicitly discovering clusters according to the most probable class, hence, the discovery of novel classes. Once we learn to recognize novel classes, we can generate pseudo-labels for novel class samples. This subsequently allows us to transform the open-world SSL problem into a closed-world SSL problem by utilizing the generated pseudo-labels of unlabeled samples to incorporate novel class samples into the labeled set. This unique perspective of transforming the open-world problem into a closed-world one is particularly powerful since it allows us to leverage any off-the-shelf closed-world SSL method to achieve further improvements. However, one shortcoming of this strategy is that the generated pseudo-labels for novel classes tend to be noisy, which can in turn impede the subsequent training. To address this issue, we introduce \emph{iterative pseudo-labeling}, a simple and efficient way to handle the noisy estimation of pseudo-labels.

\sloppy
In summary, our key contributions are: \txt{(1)} we propose a novel algorithm, OpenLDN, to solve open-world SSL. OpenLDN applies a bi-level optimization rule to determine pairwise similarity relationship without relying on pretrained features, \txt{(2)} we propose to transform the open-world SSL into a closed-world SSL problem by discovering novel classes; this allows us to leverage any off-the-shelf closed-world SSL method to further improve performance, and \txt{(3)} we introduce \emph{iterative pseudo-labeling}, a simple and efficient method to handle noisy pseudo-labels of novel classes, \txt{(4)} our experiments show that OpenLDN outperforms the existing state-of-the-art methods by a significant margin.

\vspace{-2mm}
\section{Related Works}
\vspace{-2mm}
\paragraph{\textbf{Semi-Supervised Learning:}}
SSL is a popular approach to handle label annotation bottleneck in supervised learning \cite{Gammerman1998Learning,joachims1999transductive,liu2019deep,kingma2014semi,pu2016variational,chen2020big,caron2020unsupervised}. Generally, these methods are developed for a closed-world setup, where unlabeled set only contains samples from the known classes. The two most dominant approaches for closed-world SSL are consistency regularization \cite{NIPS2016_6333,LaineA17,Miyato2018VirtualAT,NIPS2017_6719_meanT} and pseudo-labeling \cite{Lee2013PseudoLabelT,Shi_2018_ECCV,arazo2020pseudo,rizve2021in}. The consistency regularization based methods minimize a consistency loss between differently augmented versions of an image to extract salient features from the unlabeled samples. Pseudo-labeling based methods generate pseudo-labels for the unlabeled samples by a network trained on labeled data, and subsequently training on them in a supervised manner. Finally, the hybrid approaches \cite{NIPS2019_8749_MixMatch,Berthelot2020ReMixMatch:,FixMatch} combine both consistency regularization and pseudo-labeling.

Recent works \cite{oliver2018realistic,chen2020semi} demonstrate that the presence of novel class samples in the unlabeled set negatively impacts the performance on known classes. Different solutions have been proposed to address this issue \cite{guo2020safe,chen2020semi,zhao2020robust}. A weight function is trained in \cite{guo2020safe} to down-weight the novel class samples. Novel class samples are filtered out in \cite{chen2020semi} based on confidence scores. Weighted batch normalization is introduced in \cite{zhao2020robust} to achieve robustness against novel class samples. However, none of these methods attempt to solve the challenging open-world SSL problem, where the objective is to detect samples of novel classes and classify them. To the best of our knowledge, ORCA \cite{cao2022openworld} is the only method that addresses this issue by introducing an uncertainty-aware adaptive margin based cross-entropy loss to mitigate excessive influence of known classes at early stages of training. However, to discover novel classes, ORCA relies on self-supervised pretraining, which is computationally costly. To overcome the reliance on self-supervised pretraining, the pairwise similarity loss in OpenLDN exploits the information available in labeled examples from known classes using a bi-level optimization rule.

\vspace{-2mm}
\paragraph{\textbf{Novel Class Discovery:}}
Novel class discovery problem \cite{han2019learning,Han2020Automatically,hsu2018multiclass,hsu2018learning,fini2021unified,zhong2021openmix,zhao2021novel,zhong2021neighborhood,jia2021joint} is closely related to unsupervised clustering \cite{yang2016joint,yang2017towards,xie2016unsupervised,van2020scan}. The key difference between the novel class discovery and unsupervised clustering is that the former relies on an extra labeled set to learn the novel classes. To discover novel classes, \cite{Han2020Automatically} performs self-supervised pretraining followed by solving a pairwise similarity prediction task based on the rank statistics of the self-supervised features. \cite{han2019learning} extends the deep clustering framework to discover novel classes. Pairwise similarity prediction task is also applied in   \cite{hsu2018multiclass,hsu2018learning} to categorize novel classes by transferring knowledge from the known classes. While novel class discovery methods generally use multiple objective functions, \cite{fini2021unified} simplifies this using multi-view pseudo-labeling and training with cross-entropy loss. The key difference between open-world SSL and novel class discovery is that the former does not assume that unlabeled data only contains novel class samples. Hence, novel class discovery methods are not readily applicable to open-world SSL problem. Additionally, our experimentation shows that OpenLDN outperforms the appropriately modified novel class discovery methods for open-world SSL by a considerable margin.   

\vspace{-2mm}
\section{Method}
\vspace{-2mm}
\label{sec:method}

\begin{figure*}[t]
\vspace{2mm}
\begin{center}
  \includegraphics[width=\linewidth]{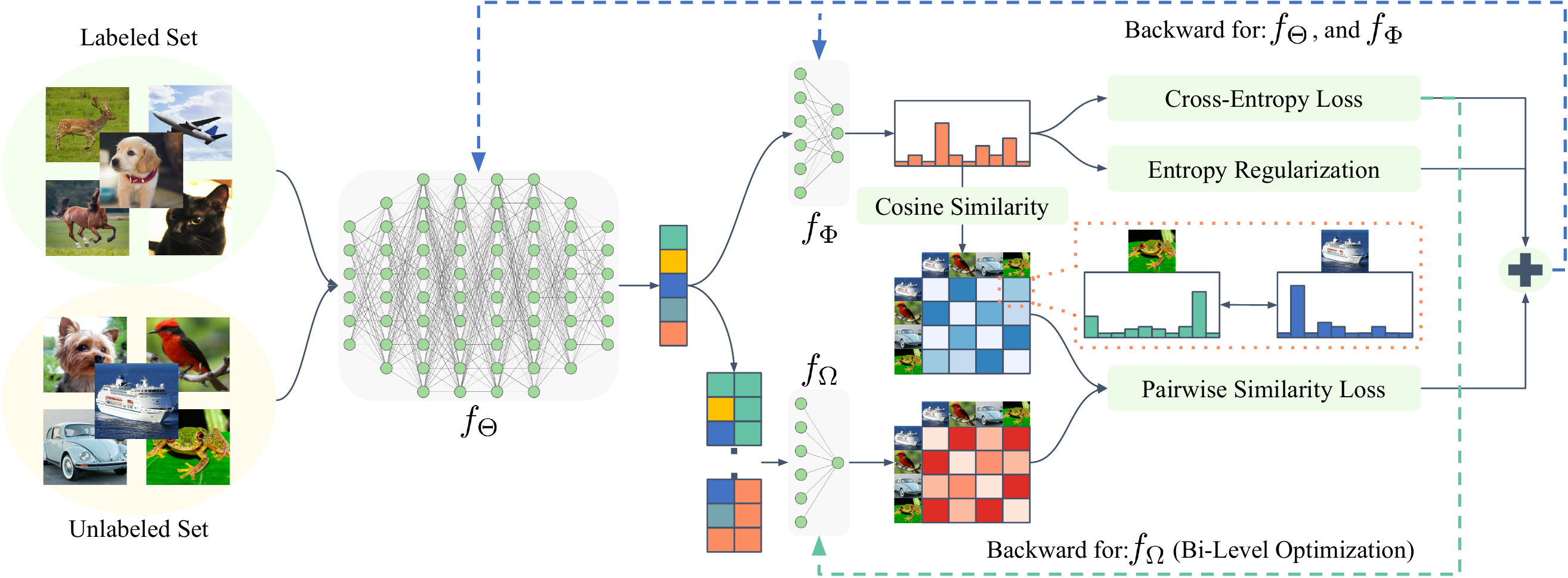}
\end{center}
\vspace{-2mm}
\caption{\small\emph{OpenLDN Overview - Learning to Discover Novel classes:} A set of labeled and unlabeled images are provided to the feature extractor, $f_\Theta$, to obtain feature embeddings. The embeddings are passed to the classifier, $f_\Phi$, to obtain output probabilities. We compute pairwise cosine similarity scores from the output probabilities for every possible pair in a batch. In parallel, the pairwise similarity prediction network, $f_\Omega$, also outputs similarity scores based on pairs of feature embeddings. Afterwards, we calculate pairwise similarity loss (Eq.~\ref{eqn:pair}) to promote the discovery of novel classes. We also compute cross-entropy (CE) loss (Eq.~\ref{eqn:ce}) and entropy regularization loss (Eq.~\ref{eqn:reg}) to complement the pairwise similarity loss by learning from labeled and pseudo-labeled samples and avoiding trivial solutions, respectively. Next, we update the parameters of $f_\Theta$ and $f_\Phi$ to minimize the overall loss. Then we compute CE loss using only the labeled samples with updated $f_\Theta$ and $f_\Phi$. Finally, we utilize a bi-level optimization rule to update $f_\Omega$ based on this CE loss (Eq.~\ref{eqn:second}). The bi-level optimization rule helps to optimize $f_\Omega$ by transferring feature similarities from known to unknown classes.}
\vspace{-2mm}
\label{fig:arch}
\end{figure*}

To identify unlabeled samples from both known and novel classes, we introduce a pairwise similarity loss to implicitly cluster the unlabeled data into known and novel classes. This implicit clustering induces discovery of novel classes which is complemented by a cross-entropy loss and an entropy regularization term. Next, we generate pseudo-labels for novel class samples to transform the original open-world SSL problem into a closed-world SSL problem. This transformation allows us to take benefit of existing off-the-shelf closed-world SSL methods to learn on both known and novel classes, delivering further gains. 
An overview of our approach is provided in Fig.~\ref{fig:arch}. In the following, we present the problem formulation and provide the details of our approach.

% \vspace{-2mm}
\subsection{Problem Formulation}
\vspace{-2mm}
We denote scalars as $a$, vectors as $\mathbf{a}$, matrices as $\mathbf{A}$, and sets as $\mathbb{A}$. In a matrix, the first index always represents rows and the second index represents columns. Further, $\mathbf{A}_{i,:}$ and $\mathbf{A}_{:,k}$ refer to the $i^{th}$ row and  $k^{th}$ column in $\mathbf{A}$, respectively.

In open-world SSL problem we assume that there is a labeled set, $\mathbb{S}_L$, and an unlabeled set, $\mathbb{S}_U$. Let, $\mathbb{S}_L = \{\mathbf{x}^l_i, {\mathbf{y}}^l_i\}_{i=1}^{n_l}$ represent the labeled dataset with $n_l$ samples, where $\mathbf{x}_i^l$ is a labeled sample and ${\mathbf{y}}^l_i$ is its corresponding label, which belongs to one of the $c_l$ known classes. Similarly, $\mathbb{S}_U=\{\mathbf{x}_{i}^u\}_{i=1}^{n_u}$, consists of $n_u$ unlabeled samples, where $\mathbf{x}_{i}^u$ belongs to one of the $c_u$ classes, where $c_u$ is the total number of classes in $\mathbb{S}_U$. In conventional closed-world SSL setting, it is assumed that the class categories for both labeled and unlabeled data are the same. However, in open-world SSL framework, $\mathbb{S}_U$ contains some samples that do not belong to any of the known classes. The samples belonging to unknown classes are called novel class samples where each sample belongs to one of the $c_n$ novel classes, i.e., in open-world setting $c_u = c_l + c_n$.  

\vspace{-4mm}
\subsection{Learning to Discover Novel Classes}
% \vspace{-2mm}
To discover novel classes OpenLDN leverages a neural network, $f_\Theta$, parameterized with $\Theta$, as feature extractor. The feature extractor generates a feature embedding by projecting an input image $\mathbf{x}$ into the embedding space, $\mathbf{z}\in\mathbb{R}^d$, i.e., $f_\Theta:\mathbb{X}\mapsto\mathbb{Z}$. Here, $\mathbb{X}$, and $\mathbb{Z}$ are sets of input images and feature embeddings, respectively. Next, to recognise samples from novel classes, as well as to classify the samples from known classes, we apply a classifier, $f_\Phi$, parameterized with $\Phi$. This classifier projects the embedding vector $\mathbf{z}$ into an output classification space,  $f_\Phi: \mathbb{Z}\mapsto \mathbb{R}^{{c_l} + {c_n}}$. In this output space, the first $c_l$ logits correspond to the known classes, and the remaining $c_n$ logits belong to novel classes. The softmax probability scores, $\mathbf{\hat{y}} \in \mathbb{R}^{{c_l} + {c_n}}$, are obtained from these output scores using softmax activation function, i.e., $\mathbf{\hat{y}}=\mathrm{Softmax}(f_{\Phi}\circ f_{\Theta}(\mathbf{x}))$.

Our overall objective to discover novel classes while recognizing known classes consists of three losses: \txt{a)} a pairwise similarity loss $\mathcal{L}_{pair}$,  \txt{b)} a cross-entropy (CE) loss $\mathcal{L}_{ce}$, and  \txt{c)} an entropy regularization term $\mathcal{L}_{reg}$. The pairwise similarity loss helps the network to discover novel classes, whereas the CE loss helps classify known classes and novel classes by utilizing the groundtruth labels and generated pseudo-labels, while the entropy regularization helps in avoiding tirvial solutions. The overall objective function to discover novel classes is as follows:
\setlength{\abovedisplayskip}{0pt}
\setlength{\belowdisplayskip}{2pt}
\setlength{\abovedisplayshortskip}{0pt}
\setlength{\belowdisplayshortskip}{0pt}
\begin{align}
\label{eqn:overall}
\vspace{-1mm}
    \mathcal{L}_{nov} = \mathcal{L}_{pair} + \mathcal{L}_{ce} + \mathcal{L}_{reg}. 
\end{align}

After training with $\mathcal{L}_{nov}$, samples that are assigned to any of the last $c_n$ logits are considered novel class samples.

\vspace{-2mm}
\paragraph{\textbf{Pairwise Similarity Loss:}}
\label{par:pair}
Discovering novel classes is a core component of our proposed method, which is an unsupervised clustering problem, that can be expressed as a pairwise similarity prediction task \cite{chang2017deep,wu2019deep}. In particular, there can only be two possible relationships between a pair of images with respect to clusters, either they belong to the same cluster or not. However, to solve a pairwise similarity prediction task supervision is needed. Previous methods \cite{cao2022openworld,Han2020Automatically} try to overcome this problem by generating pairwise pseudo-labels for all pairs of images by finding the nearest neighbors (labeled as members of the same cluster) based on pretrained features. However, such an approach is computationally expensive and suffers from noisy estimation of nearest neighbors.

In sharp contrast to this approach, instead of relying on unsupervised/self-supervised pretraining to obtain labels for the pairwise similarity prediction task, we learn to estimate the pairwise similarity scores based on the available groundtruth annotations which are more reliable. To this end, we introduce a pairwise similarity prediction network, $f_\Omega$, parameterized by $\Omega$. Given a pair of embedding vectors, $f_\Omega$ outputs a pairwise similarity score, i.e., $f_\Omega: \mathbb{Z}\times\mathbb{Z}\mapsto[0, 1]$. The pairwise similarity score from $f_\Omega$ can be used as the supervisory signal for minimizing the pairwise similarity loss. To this end, given a batch of images, we compute the cosine similarity of output probabilities between all pairs of images. After that, for our pairwise similarity loss, we minimize $l_2$ loss between the computed cosine similarity scores of output probabilities and the estimated pairwise similarity scores from $f_\Omega$. Note that minimizing pairwise similarity loss for cosine similarities of output probabilities is crucial since this will implicitly lead to formation of clusters based on maximum probability scores, consequently, the recognition of novel classes. The pairwise similarity loss is as follows:

\vspace{-2mm}
\begin{align}
\label{eqn:pair}
    \mathcal{L}_{pair} = \sum_{i\neq j}\big(\mathrm{Sim}(\mathbf{\hat{Y}}_{i,:},\mathbf{\hat{Y}}_{j,:}) - f_{\Omega}(\mathbf{Z}_{i,:}, \mathbf{Z}_{j,:})\big)^2,
\end{align}
where, $\mathbf{\hat{Y}}$ is the output probability matrix, $\mathbf{{Z}}$ is the matrix of feature embeddings, $\mathrm{Sim}(.,.)$ denotes the cosine similarity function.

To optimize the parameters of $f_\Omega$, we devise a bi-level optimization procedure \cite{bard2013practical}. Since we do not have access to the labels of any unlabeled sample, especially the samples from the novel classes, we utilize the groundtruth labels of the labeled examples belonging to known classes. The main motivation behind this bi-level optimization is to acquire a set of parameters $\Omega$ that do not deteriorate the performance of $f_\Phi \circ f_{\Theta}$ on known classes. Thereby, we optimize $f_\Omega$ based on the cross-entropy loss computed on labeled examples. The optimization procedure is as follows:

First, we update the parameters of feature extractor and classifier with the combined loss introduced in Eq.~\ref{eqn:overall} to discover novel classes.
\setlength{\abovedisplayskip}{4pt}
\setlength{\belowdisplayskip}{2pt}
\setlength{\abovedisplayshortskip}{0pt}
\setlength{\belowdisplayshortskip}{0pt}
\begin{align}
\label{eqn:first}
({\Theta}^*, {\Phi}^*) = ({\Theta}, {\Phi}) - \alpha_{(\Theta,\Phi)}\nabla_{({\Theta}, {\Phi})}\mathcal{L}_{nov}({\Theta}, {\Phi}, {\Omega}).
\end{align}
where, $\alpha_{({\Theta}, {\Phi})}$ refers to the learning rate for optimizing the parameters $\Theta$ and $\Phi$. 

Next, we use the supervised cross-entropy loss, $\mathcal{L}_{ce}^l=-\sum_i\sum_k\mathbf{Y}_{i,k}\log\mathbf{\hat{Y}}_{i,k}$, computed over the labeled examples to update the parameters of $f_\Omega$. Here, $\mathbf{Y}$ is the matrix of groundtruth labels. The update rule is according to,
\setlength{\abovedisplayskip}{4pt}
\setlength{\belowdisplayskip}{2pt}
\setlength{\abovedisplayshortskip}{0pt}
\setlength{\belowdisplayshortskip}{0pt}
\begin{align}
\label{eqn:second}
{\Omega}^* = \Omega - \alpha_{\Omega}\nabla_{\Omega}\mathcal{L}_{ce}^l({\Theta}^*, {\Phi}^*), 
\end{align}
where, $\alpha_{\Omega}$ is the learning rate for optimizing the parameters $\Omega$. Since $\Omega$ is not explicit in the objective $\mathcal{L}_{ce}^l({\Theta}^*, {\Phi}^*)$ in Eq.~\ref{eqn:second}, we perform a bi-level optimization to calculate $\nabla_{\Omega}\mathcal{L}_{ce}^l({\Theta}^*, {\Phi}^*)$. This nested optimization is available in most modern deep learning packages that support automatic differentiation \cite{paszke2017automatic,tensorflow2015}.

This bi-level optimization procedure ensures that the parameters of $f_\Omega$ are updated in such a way that the classification performance on known classes does not deteriorate since this is one of the primary objectives in open-world SSL. 

\vspace{-4mm}
\paragraph{\textbf{Learning with Labeled and Pseudo-Labeled Data:}}
In the above, we introduced the pairwise similarity loss to recognise novel classes by solving a pairwise similarity prediction task. Recall that our aim is to recognise novel classes in the unlabeled set and to classify the known classes, at the same time. This problem only allows access to a limited amount of annotations for known classes. The straightforward way to utilize these available annotations would be to minimize a cross-entropy loss on the labeled samples. However, this approach can create a strong bias towards known classes because of their strong training signal \cite{cao2022openworld}. To mitigate this bias and to utilize the unlabeled samples more efficiently, we generate pseudo-labels for all the unlabeled data. The generated pseudo-labels can be used with groundtruth labels for minimizing cross-entropy loss.

Following the common practice \cite{Lee2013PseudoLabelT,arazo2020pseudo,rizve2021in}, we generate pseudo-labels based on the network output probabilities. To reduce the possibility of erroneous training with unreliable pseudo-labels, we generate pseudo-labels only for sufficiently confident predictions. In addition, our pseudo-label based cross-entropy learning satisfies another commonly used objective in SSL works i.e., consistency regularization. This objective encourages perturbation invariant output distribution so that the decision boundaries lie in low density regions \cite{chapelle2005semi,verma2019interpolation}. One way to satisfy this objective is to minimize the divergence between output probabilities of two randomly transformed versions of an image. However, it adds another term to the loss, and consequently a new hyperparameter. A more elegant way is to use the generated pseudo-label from one transformed version of an image as the target for the other version. We utilize the pseudo-labels generated from a weakly transformed version of an image $\mathbf{x}^w$ as the target for its strongly augmented version, $\mathbf{x}^s$. We state our pseudo-label generation process below:

\setlength{\abovedisplayskip}{-4pt}
\setlength{\belowdisplayskip}{4pt}
\setlength{\abovedisplayshortskip}{0pt}
\setlength{\belowdisplayshortskip}{0pt}
\begin{align}
    \mathbb{S}_{PL} =\{(\mathbf{x}_{i}^s, \mathbbm{1}_{\max(\mathbf{\hat{Y}}_{i,:}^w)}(\mathbf{\hat{Y}}_{i,:}^w)|\max(\mathbf{\hat{Y}}_{i,:}^w)>\tau\},
\end{align}
\noindent where, $\tau=0.5$ (midpoint in binary classification) to avoid per-dataset finetuning.

Once the pseudo-labels are generated we combine them with the groundtruth labels, $\mathbb{S}=\mathbb{S}_{PL}\cup\mathbb{S}_{L}$, and train the model using cross-entropy loss. In practice, we combine these two sets within a batch. Let $\mathbb{S}_B$ denote a batch, the cross-entropy loss on this set is defined as:

\setlength{\abovedisplayskip}{-8pt}
\setlength{\belowdisplayskip}{2pt}
\setlength{\abovedisplayshortskip}{0pt}
\setlength{\belowdisplayshortskip}{0pt}
\begin{align}
\label{eqn:ce}
\vspace{-2mm}
   \mathcal{L}_{ce} = -\sum_{i\in\mathbb{S}_B}\sum_{k=1}^{c_u}\mathbf{\tilde{Y}}_{i,k}\log\mathbf{\hat{Y}}_{i,k},
\end{align}
where, $\mathbf{\tilde{Y}}$ is the matrix of one-hot encoded ground-truth labels and the generated pseudo-labels.

\vspace{-4mm}
\paragraph{\textbf{Entropy Regularization:}}
\label{par:entropy}
One of the well-known drawbacks of assigning unlabeled data to distinct categories based on a discriminative (such as cross-entropy) loss is that it can lead to a trivial solution where all the unlabeled samples are assigned to the same class \cite{YM.2020Self-labelling,caron2018deep,cao2022openworld,fini2021unified}. Our pairwise similarity loss suffers from the same problem since such a solution will also minimize our pairwise similarity loss in Eq.~\ref{eqn:pair}. To address this, we incorporate an entropy regularization term in our training objective. One way to achieve this would be to apply entropy regularization to the output of each sample independently. However, this way of entropy minimization leads to substantial changes in the individual output probabilities which results in arbitrary class assignments for novel class samples. To avoid this problem, we apply entropy regularization over an aggregated statistic, in our case, the mean of the sample probabilities of an entire batch. This entropy regularization term prevents a single class from dominating the entire batch, where most of unlabeled samples are only assigned to one class. %, 
This term does not interfere with the balanced class assignments. 
The entropy regularization is defined as, 

\begin{align}
    \label{eqn:reg}
    % \mathbf{\Bar{y}} = \frac{1}{B} \sum_{i=1}^{B}\mathbf{\hat{y}}^{(i)}\notag\\
    \mathcal{L}_{reg} = \sum_{k=1}^{c_u}\mathbf{\Bar{y}}_{k}\log\mathbf{\Bar{y}}_{k},
\end{align}
where, $\mathbf{\Bar{y}} = \frac{1}{b} \sum_{i=1}^{b}\mathbf{\hat{Y}}_{i,:}$ is the average probability of the batch, and $b$ denotes the number of examples in a batch.

\vspace{-2mm}
\subsection{Closed-World Training with Iterative Pseudo-Labeling}
\label{sec:ipl}
\vspace{-2mm}
 
Once we discover novel classes in the unlabeled data we can reformulate the open-world SSL problem as a closed-world one to improve performance. To this end, we generate the pseudo-labels for all novel class samples using Eq.~\ref{eqn:pl}:

\begin{align}
\label{eqn:pl}
    \mathbf{\check{Y}} = \mathbbm{1}_{\max(\mathbf{\hat{Y}}_{i,:})}(\mathbf{\hat{Y}}_{i,:}).
\end{align}

Next, using generated pseudo-labels, we add novel class samples to the labeled set. At this point, we are able to apply any standard closed-world SSL method \cite{NIPS2019_8749_MixMatch,xie2019unsupervised,FixMatch,NIPS2017_6719_meanT}. Unfortunately, pseudo-labels tend to contain noise that can hamper the performance. To mitigate the negative impact of noise, we propose to perform pseudo-labeling during the closed-world SSL training in an iterative manner. This new iterative pseudo-labeling approach can be related to EM algorithm. From this perspective, we iteratively attempt to update the pseudo-labels (expectation step), and train the network by minimizing the loss on those updated pseudo-labels (maximization step). It is important to note that OpenLDN, including the final closed-world SSL retraining, is computationally lighter or comparable to the unsupervised/self-supervised pretraining based approaches (Sec.~\ref{par:time_complexity}). Besides, the transformation from the open-world SSL problem to a closed-world problem is a general solution which can be applied to other methods as well. We provide our overall training algorithm in supplementary materials.

%%%%%%%%%%%%%%%%%%%%%%%%%%%%%%%%%%%%%%%%%%%%%%%%%%%%%%%%%%%%
\vspace{-2mm}
\section{Experimental Evaluation}
\vspace{-2mm}
\label{sec:exp}
\paragraph{\textbf{Datasets:}}
\label{para:dataset}
To demonstrate the effectiveness of OpenLDN, we conduct experiments on five common benchmark datasets: CIFAR-10 \cite{cifar10}, CIFAR-100 \cite{cifar100}, ImageNet-100\cite{deng2009imagenet}, Tiny ImageNet \cite{le2015tiny}, and Oxford-IIIT Pet dataset \cite{parkhi12a}. Both CIFAR-10 and CIFAR-100 datasets contain 60K images (split into 50K/10k train/test set), and they have 10 and 100 categories, respectively. ImagNet1-100 dataset contains 100 image categories from ImageNet. Tiny ImageNet contains 100K/10K training/validation images from 200 classes. Finally, Oxford-IIIT Pet contains images from 37 categories split into 3680/3669 train/test set. In our experiments, we divide each of these datasets based on the percentage of known and novel classes. We consider the first $c_l$ classes as known and the rest as novel. For known classes, we randomly select a portion of data to construct the labeled set and add the rest to the unlabeled set along with all novel class samples.

 \vspace{-2mm}
\paragraph{\textbf{Implementation Details:}}
\label{para:implimentation}
We use ResNet-18 \cite{he2016deep} as the feature extractor in all of our experiments except the experiments on ImageNet-100 dataset where we use ResNet-50. We instantiate our pairwise similarity prediction network, $f_{\Omega}$, with an MLP consisting of a single hidden layer of dimension 100. The classifier, $f_{\Phi}$, is a single linear layer. To discover novel classes, we train for 50 epochs with a batch size of 200 (480 for ImageNet-100) in all the experiments. We always use Adam optimizer \cite{kingma2014adam}. For training the feature extractor and the classifier, we set the learning rate to $5e^{-4}$ ($1e^{-2}$ for ImageNet-100). For the pairwise similarity prediction network, we use a learning rate of $1e^{-4}$. We use two popular closed-world SSL methods, Mixmatch\cite{NIPS2019_8749_MixMatch} and UDA\cite{xie2019unsupervised}, for second stage closed-world SSL training. For this closed-world training, to preserve data balance, we select an equal number of pseudo-labels for each novel class. For iterative pseudo-labeling, we generate pseudo-labels every 10 epochs. Additional implementation details are available in the supplementary materials.

\vspace{-2mm}
\paragraph{\textbf{Evaluation Metrics:}}
We report standard accuracy for known classes. In addition, following \cite{han2019learning,Han2020Automatically,cao2022openworld,fini2021unified}, we report clustering accuracy on novel classes. %To this end, 
We leverage the Hungarian algorithm \cite{kuhn1955hungarian} to align the predictions and groundtruth labels before measuring the classification accuracy. Finally, we also report the joint accuracy on the novel and known classes by using the Hungarian algorithm.
\vspace{-2mm}

%%%%%%%%%%%%%%%%%%%%%%%%%%%%%%%%%%%%%%%%%%%%%%%%%%%%%%%%%%%%%%%%%%%%%%%%%%%%%%%%%%%%%%%
\begin{table}[t]
\vspace{2mm}
\begin{center}\setlength{\tabcolsep}{2pt}
\small
\resizebox{\textwidth}{!}{%
\begin{tabular}{lccccccccc}
\hline

\hline

\hline\\[-3mm]
 \multicolumn{1}{c}{\multirow{2}{*}{\textbf{Method}}} &
 \multicolumn{3}{c}{\textbf{CIFAR10}} &
 \multicolumn{3}{c}{\textbf{CIFAR100}} &
 \multicolumn{3}{c}{\textbf{ImageNet100}} \\  
\multicolumn{1}{c}{} & \textbf{Known} & \textbf{Novel} & \textbf{All} & \textbf{Known} & \textbf{Novel} & \textbf{All} & \textbf{Known} & \textbf{Novel} & \textbf{All}
 \\[-3mm]
\\
 \hline

\hline

\hline
FixMatch\cite{FixMatch} & $71.5$ & $50.4$ & $49.5$ & $39.6$ & $23.5$ & $20.3$ & 65.8 & 36.7 & 34.9\\
DS$^{3}$L\cite{guo2020safe} & $77.6$ & $45.3$ & $40.2$ & $55.1$ & $23.7$ & $24.0$ & 71.2 & 32.5 & 30.8\\
CGDL\cite{sun2020conditional} & $72.3$ & $44.6$ & $39.7$ & $49.3$ & $22.5$ & $23.5$ & 67.3 & 33.8 & 31.9\\
DTC~\cite{han2019learning} & $53.9$ & $39.5$ & $38.3$ & $31.3$ & $22.9$ & $18.3$ & 25.6 & 20.8 & 21.3\\
RankStats\cite{Han2020Automatically} & $86.6$ & $81.0$ & $82.9$ & $36.4$ & $28.4$ & $23.1$ & 47.3 & 28.7 & 40.3\\
UNO\cite{fini2021unified} & $91.6$ & $69.3$ & $80.5$ & $68.3$ & $36.5$ & $51.5$ & $-$ & $-$ & $-$\\
ORCA\cite{cao2022openworld} & $88.2$ & $90.4$ & $89.7$ & $66.9$ & $43.0$ & $48.1$ & 89.1 & $\mathbf{72.1}$ & 77.8\\
\rowcolor[gray]{.95} {OpenLDN-MixMatch} & $95.2$ & $92.7$ & $94.0$ & $73.5$ & $\mathbf{46.8}${\tiny{\textcolor{teal}{$\mathord{\uparrow}3.8$}}} & $\mathbf{60.1}${\tiny{\textcolor{teal}{$\mathord{\uparrow}8.6$}}} & $-$ & $-$ & $-$\\ 
 
\rowcolor[gray]{.95} {OpenLDN-UDA} & $\mathbf{95.7}${\tiny{\textcolor{teal}{$\mathord{\uparrow}4.1$}}} & $\mathbf{95.1}${\tiny{\textcolor{teal}{$\mathord{\uparrow}4.7$}}} & $\mathbf{95.4}${\tiny{\textcolor{teal}{$\mathord{\uparrow}5.7$}}} & $\mathbf{74.1}${\tiny{\textcolor{teal}{$\mathord{\uparrow}5.8$}}} & $44.5$ & $59.3$ & $\mathbf{89.6}${\tiny{\textcolor{teal}{$\mathord{\uparrow}0.5$}}} & $68.6${\tiny{\textcolor{teal}{$\mathord{\downarrow}3.5$}}} & $\mathbf{79.1}${\tiny{\textcolor{teal}{$\mathord{\uparrow}1.3$}}} 
\\\hline 

\hline

\hline
\end{tabular}
}
\end{center}
% \vspace{-6mm}
\caption{\small Accuracy on \textbf{CIFAR-10}, \textbf{CIFAR-100}, and \textbf{ImageNet-100} datasets with 50\% classes as known and 50\% classes as novel.}
\label{tab:cifar10}
\vspace{-4mm}
\end{table}
%%%%%%%%%%%%%%%%%%%%%%%%%%%%%%%%%%%%%%%%%%%%%%%%%%%%%%%%%%%%

\vspace{-2mm}
\subsection{Results}
% \vspace{-2mm}
\paragraph{\textbf{CIFAR-10, CIFAR-100, and ImageNet-100 Experiments:}}
We present our experimental results on CIFAR-10, CIFAR-100, and ImageNet-100 datasets in Tab.~\ref{tab:cifar10}. We conduct experiments with 50\% novel classes on all three datasets, where we include 50\% labeled data from known classes. We report additional results with less labeled data in supplementary material. For comparison, we primarily use the scores reported in \cite{cao2022openworld}. Furthermore, as another competitive baseline, we modify a recent novel class discovery method, UNO~\cite{fini2021unified}, and include its performance for comparison. Tab.~\ref{tab:cifar10} shows that both OpenLDN-MixMatch and OpenLDN-UDA significantly outperform novel class discovery methods (DTC~\cite{han2019learning}, RankStats~\cite{Han2020Automatically}, and UNO~\cite{fini2021unified}) that have been modified for open-world SSL task. OpenLDN also outperforms other baseline methods: FixMatch~\cite{FixMatch}, DS$^3$L~\cite{guo2020safe}, and CGDL~\cite{sun2020conditional}. These results showcase the efficacy of OpenLDN, where it outperforms previous state-of-the-art (ORCA~\cite{cao2022openworld}) by about 4.7-7.5\% absolute improvement on different evaluation metrics on CIFAR-10 dataset. We observe a similar pattern on CIFAR-100 dataset, where OpenLDN outperforms ORCA and UNO by 12\% and 8.6\% respectively on the joint task of classifying known and novel classes. We also notice a similar trend on ImageNet-100 dataset. In parallel to outperforming all the baselines methods, OpenLDN achieves a modest 1.3\% improvement over ORCA. These results validate the effectiveness of OpenLDN in solving open-world SSL problem. 

%%%%%%%%%%%%%%%%%%%%%%%%%%%%%%%%%%%%%%%%%%%%%%%%%%%%%%%%%%%%%%%%%%%%%%%%%%%%%%%%%%%
\begin{table}[t]
\vspace{2mm}
\begin{center}\setlength{\tabcolsep}{2pt}
\small
\resizebox{0.9\textwidth}{!}{%
\begin{tabular}{lcccccc}
\hline

\hline

\hline\\[-3mm]
 \multicolumn{1}{c}{\multirow{2}{*}{\textbf{Method}}} &
 \multicolumn{3}{c}{\textbf{Tiny ImageNet}} &
 \multicolumn{3}{c}{\textbf{Oxford-IIIT Pet}} \\  
\multicolumn{1}{c}{} & \textbf{Known} & \textbf{Novel} & \textbf{All} & \textbf{Known} & \textbf{Novel} & \textbf{All}
 \\[-3mm]
\\
 \hline

\hline

\hline
DTC~\cite{han2019learning} & $28.8$ & $16.3$ & $19.9$ & $20.7$ & $16.0$ & $13.5$ \\
RankStats~\cite{Han2020Automatically} &$5.7$ & $5.4$ & $3.4$ & $12.6$ & $11.9$ & $11.1$ \\
UNO~\cite{fini2021unified} &  $46.5$ & $15.7$ & $30.3$ & $49.8$ & $22.7$ & $34.9$\\ 
\rowcolor[gray]{.95} {OpenLDN-MixMatch} & $52.3$ & $19.5$ & $36.0$ & $\mathbf{67.1}${\tiny{\textcolor{teal}{$\mathord{\uparrow}17.3$}}} & $27.3$ & $47.7$\\ 
\rowcolor[gray]{.95} {OpenLDN-UDA} & $\mathbf{58.3}${\tiny{\textcolor{teal}{$\mathord{\uparrow}11.8$}}} & $\mathbf{25.5}${\tiny{\textcolor{teal}{$\mathord{\uparrow}9.8$}}} & $\mathbf{41.9}${\tiny{\textcolor{teal}{$\mathord{\uparrow}11.6$}}} & $66.8$ & $\mathbf{33.1}${\tiny{\textcolor{teal}{$\mathord{\uparrow}10.4$}}} & $\mathbf{50.4}${\tiny{\textcolor{teal}{$\mathord{\uparrow}15.5$}}}\\ \hline 

\hline

\hline
\end{tabular}
}
\end{center}
% \vspace{-6mm}
\caption{\small Accuracy on \textbf{Tiny ImageNet} and \textbf{Oxford-IIIT Pet} datasets with 50\% classes as known and 50\% classes as novel.}
\label{tab:tiny_pet}
\vspace{-8mm}
\end{table}
% %%%%%%%%%%%%%%%%%%%%%%%%%%%%%%%%%%%%%%%%%%%%%%%%%%%%%%%%%%%%%%%%%%%%%%%%%%%%%%%%%%%%%%

\vspace{-2mm}
\paragraph{\textbf{Tiny ImageNet and Oxford-IIIT Pet Experiments:}} We also conduct additional experiments on the challenging Tiny ImageNet dataset, where the total number of classes is significantly larger than CIFAR-10, CIFAR-100, and ImageNet-100 datasets. Moreover, to further demonstrate the effectiveness of OpenLDN, we also conduct experiments on a fine-grained dataset, i.e., Oxford-IIIT Pet. The results of these experiments are presented in Tab.~\ref{tab:tiny_pet}. On Tiny ImageNet dataset we observe that OpenLDN significantly outperforms DTC and RankStat. Furthermore, OpenLDN-UDA achieves $\sim$60\% relative improvement on novel classes over UNO. OpenLDN-MixMatch also achieves a significant improvement over UNO on novel and all classes. Furthermore, on fine-grained Oxford-IIIT Pet dataset, we make a similar comparison and observe that OpenLDN significantly outperforms all three novel class discovery methods by a large margin. To be precise, OpenLDN-Mixmatch achieves 12.8\% absolute improvement over UNO on the joint classification task and similarly, OpenLDN-UDA achieves 15.5\% absolute improvement. Experiments on both of these datasets demonstrate that OpenLDN can scale up to a large number of classes and is also effective for challenging fine-grained classification task.

%%%%%%%%%%%%%%%%%%%%%%%%%%%%%%%%%%%%%%%%%%%%%%%%%%%%%%%%%%%%%%%%%%%%%%%%%%%%%%%%%%%%%%
\begin{table}[t]
\vspace{2mm}
% \vspace{-6mm}
\begin{center}\setlength{\tabcolsep}{4pt}
\small
\resizebox{0.7\textwidth}{!}{%
\begin{tabular}{ccccccc}
\hline

\hline

\hline\\[-3mm]
\textbf{EntReg} & \textbf{SimLoss} & \textbf{CWT} & \textbf{ItrPL} & \textbf{Known} & \textbf{Novel} & \textbf{All}
 \\[-3mm]
\\
 \hline

\hline

\hline
\xmark & \cmark & \xmark & \xmark & $66.7$ & $-$ & $33.4$\\
\cmark & \xmark & \xmark & \xmark & $66.2$ & $26.6$ & $46.2$  \\
\cmark & \cmark & \xmark & \xmark & $66.2$ & $40.3$ & $53.3$  \\
\cmark & \cmark & \cmark & \xmark & $73.9$ & $44.9$ & $59.1$ \\ 
\cmark & \cmark & \cmark & \cmark & $73.5$ & $46.8$ & $60.1$ \\ \hline 

\hline

\hline
\end{tabular}
}
\end{center}
% \vspace{-6mm}
\caption{\small Ablation study on \textbf{CIFAR-100} with 50\% classes as known and 50\% classes as novel. Here, \textbf{EntReg} refers to entropy regularization, \textbf{SimLoss} means pairwise similarity loss, \textbf{CWT} refers to closed-world SSL training, and \textbf{ItrPL} denotes iterative pseudo-labeling. Each component of OpenLDN contributes towards final performance. }
\label{tab:ablation}
\vspace{-8mm}
\end{table}
%%%%%%%%%%%%%%%%%%%%%%%%%%%%%%%%%%%%%%%%%%%%%%%%%%%%%%%%%%%%%%%%%%%%%%%%%%%%%%%%

\vspace{-2mm}
\subsection{Ablation and Analysis}
\vspace{-2mm}

We conduct extensive ablation studies on the CIFAR-100 dataset, with 50\% labels, to study the contribution of different components of OpenLDN. The results are presented in Tab.~\ref{tab:ablation}. In this table, the first row demonstrates that without entropy regularization OpenLDN is unable to detect novel classes. We contribute this to overpowering of a single class (sec.~\ref{par:entropy}). Next, we evaluate the impact of our pairwise similarity loss with bi-level optimization rule. We observe that without pairwise similarity loss the performance of OpenLDN degrades by 13.7\% on novel classes which makes it the most critical component of our proposed solution. We also observe that our pairwise similarity loss does not sacrifice known class performance to improve the performance on novel classes. This outcome is expected since one of the objectives of our bi-level optimization rule is to retain performance on known classes (sec.~\ref{par:pair}). The fourth row demonstrates the effectiveness of transforming the open-world SSL problem into a closed-world one. Here, we observe that with this component we obtain a significant improvement in known class performance. In addition, we also notice a significant improvement in novel class performance. Interestingly, on this dataset, OpenLDN outperforms ORCA~\cite{cao2022openworld} (on joint classification task) even without the subsequent closed-world SSL training. Finally, Tab.~\ref{tab:ablation} shows that including iterative pseudo-labeling proves to be effective, where we observe $\sim$2\% performance boost on novel classes. In conclusion, this extensive ablation study empirically validates the effectiveness of different components of our solution.   

\begin{figure}[t]
\captionsetup[subfloat]{labelformat=empty}
\vspace{-2mm}
    \centering
    \subfloat[]{{\includegraphics[width=0.4\textwidth]{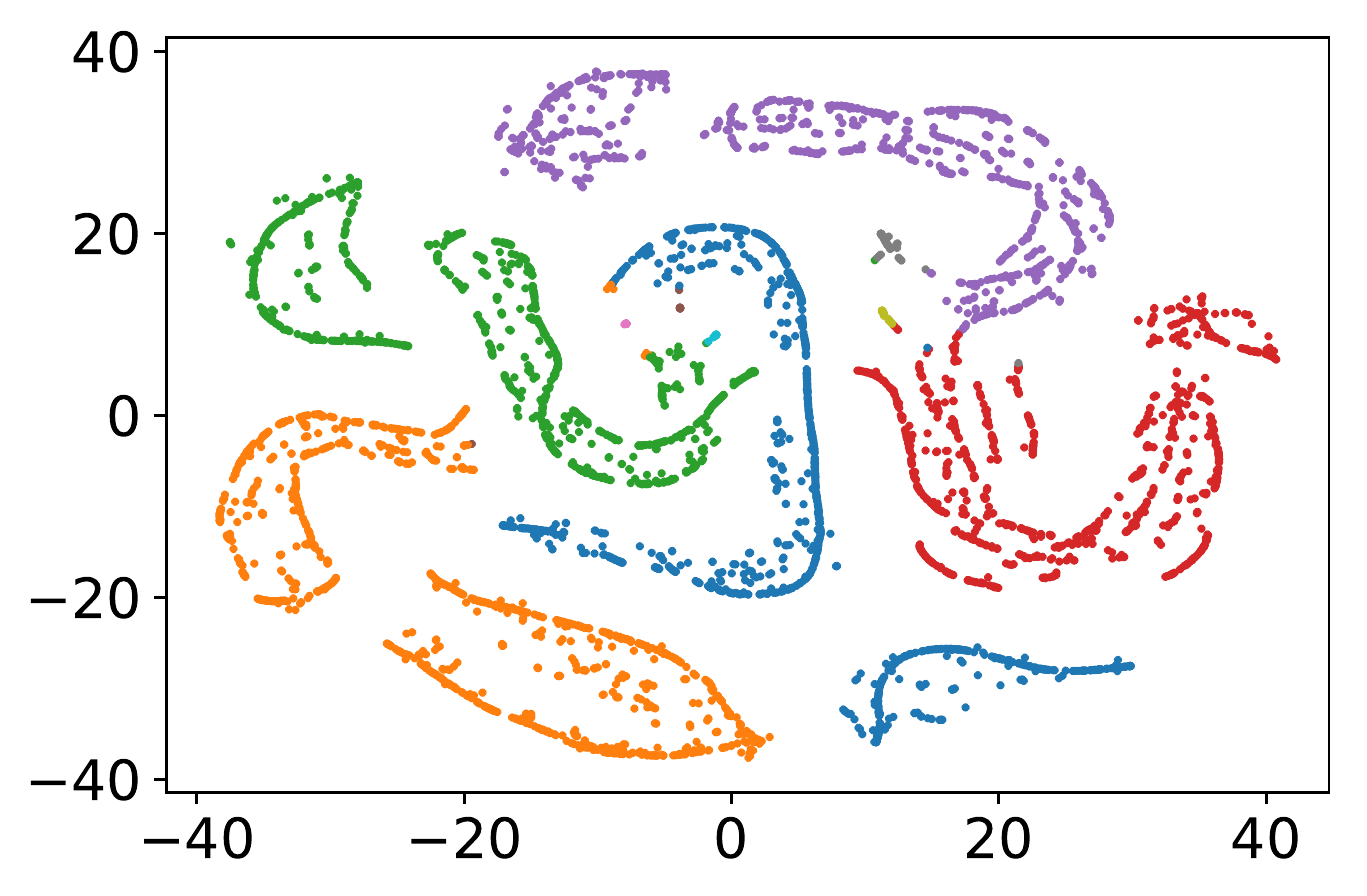}{\label{fig:ece}} }}
    \qquad
    \subfloat[]{{\includegraphics[width=0.4\textwidth]{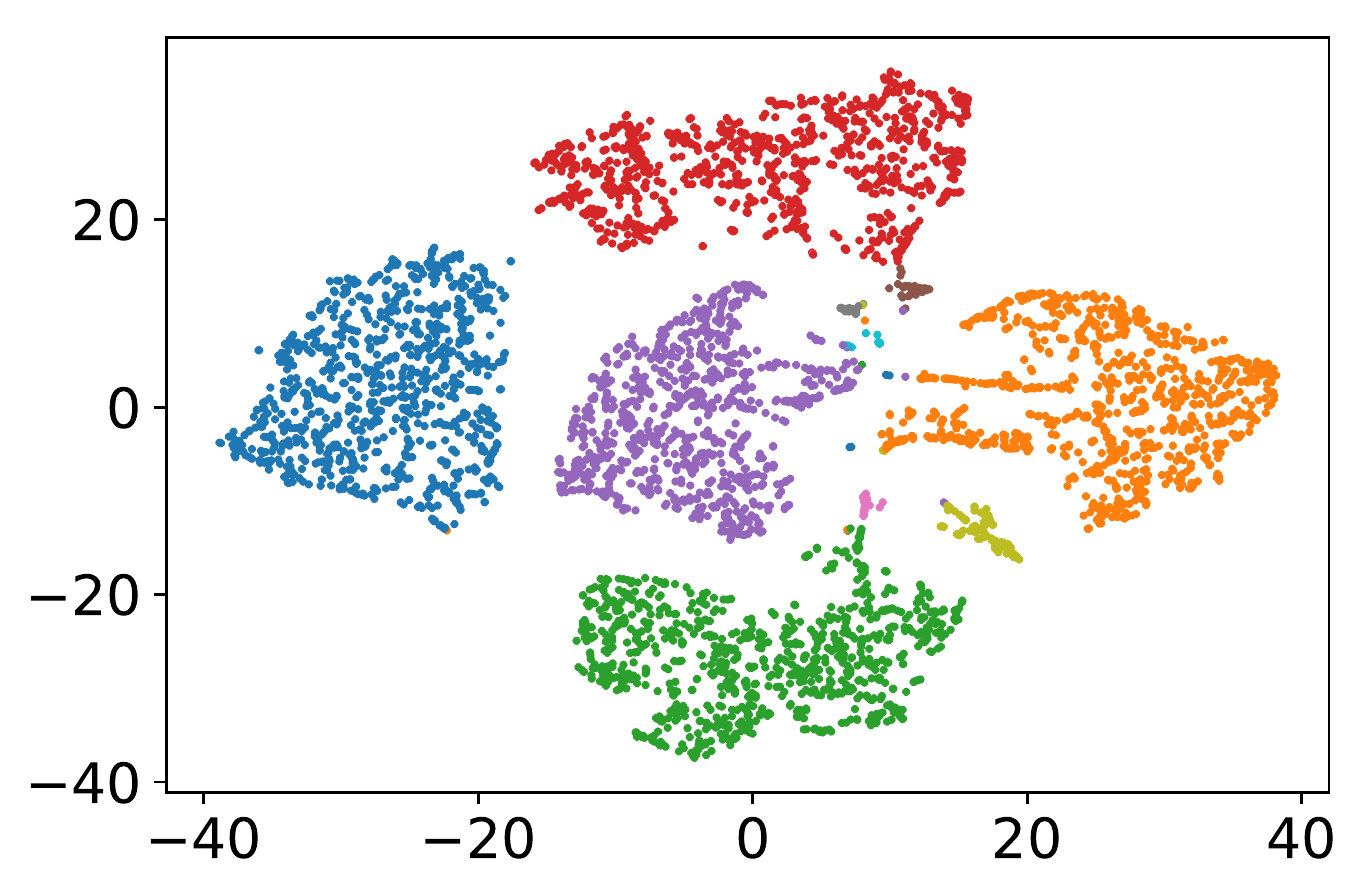}{\label{fig:acc}} }}
    % \qquad

    \vspace{-6mm}
    \caption{\small Impact of closed-world training. t-SNE visualization of novel class probabilities on \textbf{CIFAR-10}: (left to right) before and after closed-world training.  
    %(left), and after closed-world training (right).
    }
    \label{fig:tsne}%
    \vspace{-6mm}
\end{figure}

\paragraph{\textbf{Impact of Closed-World Training:}}
In an attempt to further investigate the impact of closed-world SSL training, we perform a t-SNE visualization of probability outputs of the novel class samples on CIFAR-10 dataset in Fig.~\ref{fig:tsne}. Following our general setup, in this experiment, we consider 50\% classes as novel. The results in Fig.~\ref{fig:tsne} show that after novel class discovery training the novel classes form very distinct clusters. However, since novel classes are learned through auxiliary losses, without any direct supervision, there is some overlap between different classes. After training with a closed-world SSL method (MixMatch) we observe that these overlaps fade away and novel classes become well-separated and form compact clusters. This analysis further validates the complimentary effect of incorporating a closed-world SSL method after discovering novel classes.

% 
%%%%%%%%%%%%%%%%%%%%%%%%%%%%%%%%%%%%%%%%%%%%%%%%%%%%%%%%%%%%%%%%%%%%%%%%%%%%%%%%%%%%%%
% \begin{table}[ht]
\begin{wraptable}{R}{0.5\textwidth}
% \vspace{-1.3cm}
\vspace{-10mm}
\begin{center}
\small
\resizebox{\linewidth}{!}{%
\begin{tabular}{lccc}
\hline

\hline

\hline\\[-3mm]
\textbf{Pairwise Sim. Est.} & \textbf{Known} & \textbf{Novel} & \textbf{All}
 \\[-3mm]
\\
 \hline

\hline

\hline
No similarity & $66.2$ & $26.6$ & $46.2$\\
Soft Cosine & $64.5$ & $10.3$ & $37.4$\\
Hard Cosine (0.50) & $53.7$ & $2.1$ & $27.9$  \\
Hard Cosine (0.95) & $54.2$ & $17.3$ & $35.8$  \\
Nearest Neighbor & $\mathbf{66.4}$ & $31.7$ & $49.1$ \\
% OpenLDN-Pairwise (Iterative Update) & $65.1$ & $10.5$ & $37.8$ \\
\rowcolor[gray]{.95} {OpenLDN} & $66.2$ & $\mathbf{40.3}$ & $\mathbf{53.3}$ \\ \hline 

\hline

\hline
\end{tabular}
}
\end{center}
 \vspace{-4mm}
\caption{\small Results with alternate pairwise similarity estimation methods on \textbf{CIFAR-100} dataset with 50\% classes as known and 50\% classes as novel.}
\label{tab:pairwise}
\vspace{-4mm}
% \end{table}
\end{wraptable}

%  \vspace{-4mm}
\paragraph{\textbf{Effect of Changing Pairwise Similarity Estimation:}}
\label{par:alt_loss}
In another set of experiments, to analyze the effectiveness of our pairwise similarity estimation with bi-level optimization rule we conduct experiments with alternate pairwise similarity estimation methods on CIFAR-100 dataset. The results are reported in Tab.~\ref{tab:pairwise}. In this table, to provide a baseline to compare other pairwise similarity estimation techniques, we include the performance of OpenLDN without any pairwise similarity loss in the first row. The next row in the table demonstrates the performance of OpenLDN when the pairwise similarity is directly estimated from the cosine similarity of features. Surprisingly this method of estimating pairwise similarity performs worse than our baseline without any pairwise similarity loss (first row). We hypothesize this phenomenon to instability of cosine similarity of features without any feature pretraining. In the next set of experiments (third and fourth rows), we utilize a hard version of this pairwise similarity estimation method, where we threshold the pairwise feature similarities using two different thresholds (0.50 and 0.95). After that, we minimize a binary cross-entropy loss as the pairwise similarity loss. The results of these two experiments are reported in third and fourth row, where a further drop in performance is observed. One possible explanation is that without any feature pretraining this kind of pairwise similarity estimation leads to a lot of false positives (from novel classes) which in turn reduces the score of known classes. Finally, in the next row of the table, we use a nearest neighbor based pairwise similarity estimation technique similar to ORCA \cite{cao2022openworld}. As indicated in the table, this nearest neighbor based similarity estimation improves over the baseline without any pairwise similarity loss. However, our pairwise similarity estimation based on the bi-level optimization rule outperforms this nearest neighbor based estimation technique by a significant margin; we obtain $\sim$9\% absolute improvement on novel classes. These set of experiments further validate our claim that common pairwise similarity estimation techniques are not potent without feature pretraining, whereas, the proposed pairwise similarity estimation in OpenLDN is able to learn these pairwise similarities effectively.

\begin{wrapfigure}{R}{0.5\textwidth}
% \vspace{-2.2cm}
\vspace{-12mm}
\begin{center}
  \includegraphics[width=\linewidth]{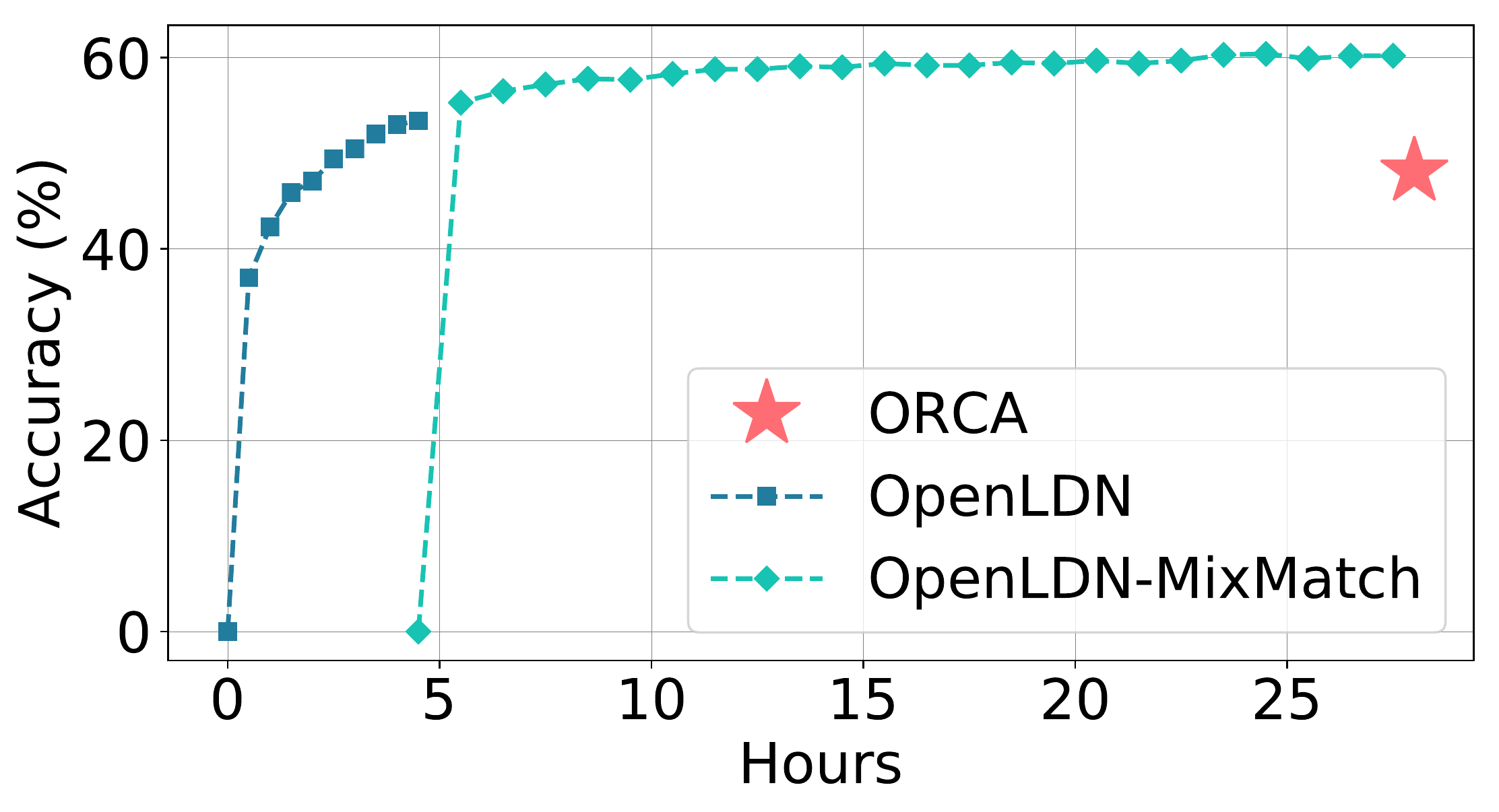}
\end{center}
\vspace{-6mm}
\caption{\small Accuracy with respect to wall-clock time on \textbf{CIFAR-100} dataset. OpenLDN outperforms ORCA in less than 3 hours.}
\vspace{-6mm}
\label{fig:time_complx}
\end{wrapfigure}

\vspace{-2mm}
\paragraph{\textbf{Computational Cost Analysis:}}
\label{par:time_complexity}
One of the primary advantages of OpenLDN is that unlike ORCA it does not require any feature initialization technique. This enables OpenLDN to be computationally efficient in comparison to ORCA. To demonstrate the efficiency of OpenLDN we report the performance on CIFAR-100 dataset across different training budgets in Fig.~\ref{fig:time_complx}. We notice that, ORCA with SimCLR pretraining takes $\sim28$ hours with our computing resources. We also observe that OpenLDN outperforms ORCA under $3$ hours on this dataset even without the closed-world SSL training. On the other hand, the closed-world SSL training reaches a reasonable performance very fast and improves relatively slowly over time. Therefore, if computational budget is of concern we can stop training at an earlier stage without making a noticeable trade-off in performance.

\begin{figure}[t]
\captionsetup[subfloat]{labelformat=empty}
% \vspace{-4mm}
    \centering
    \subfloat[]{{\includegraphics[width=0.45\textwidth]{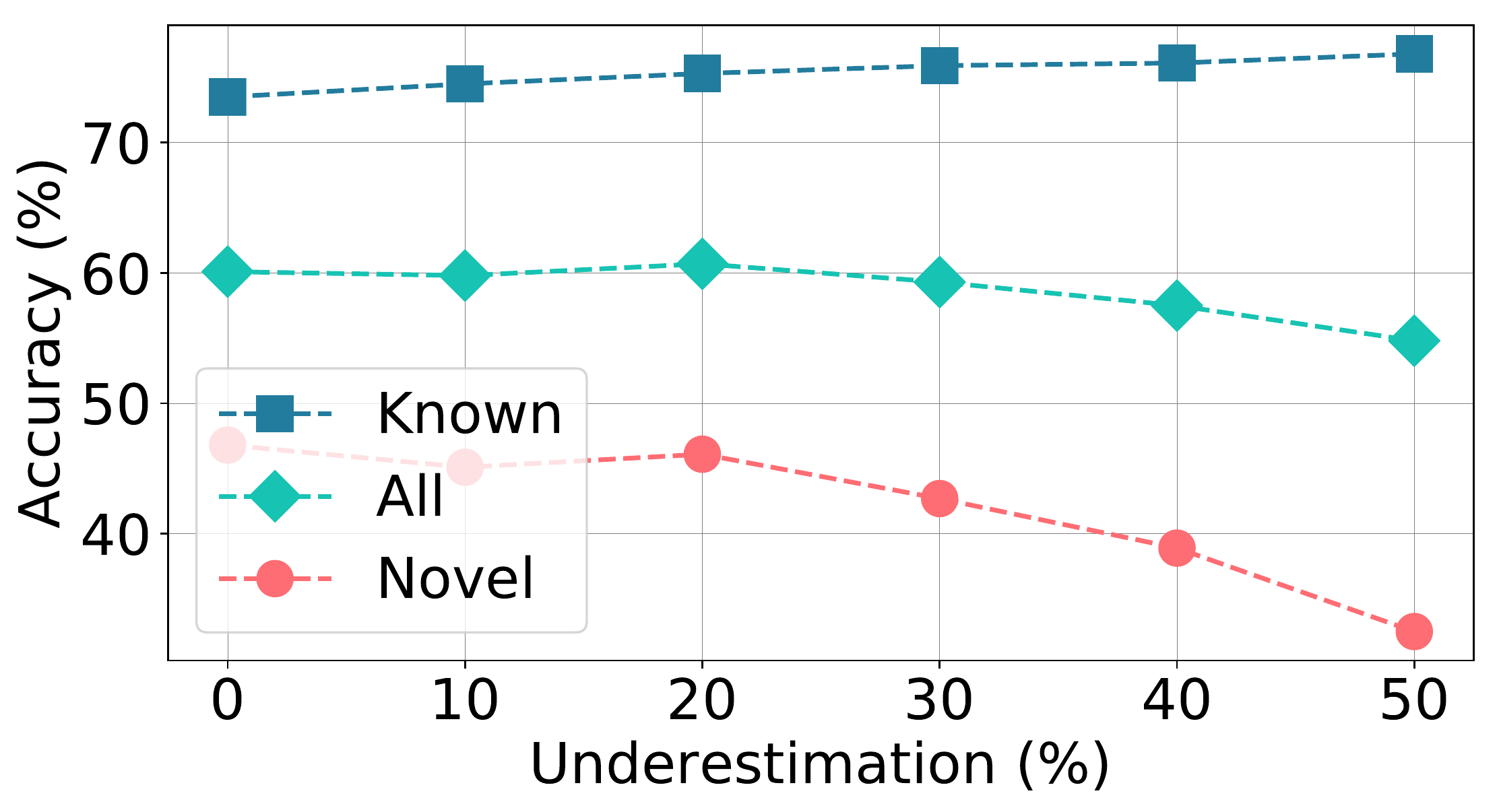}{\label{fig:under}} }}
    \qquad
    \subfloat[]{{\includegraphics[width=0.45\textwidth]{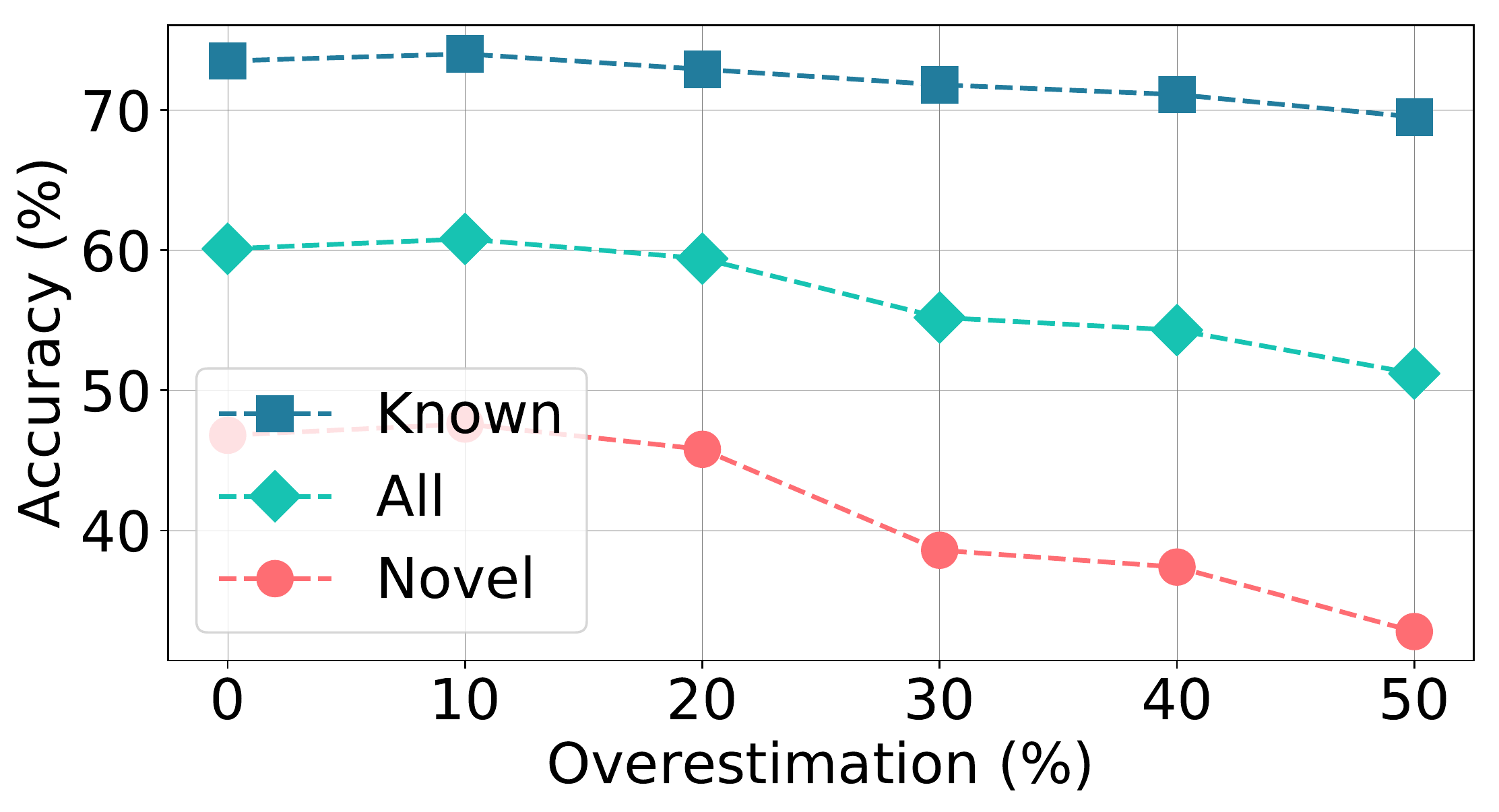}{\label{fig:over}} }}
    % \qquad

    \vspace{-8mm}
    \caption{\small Performance on \textbf{CIFAR-100} dataset with unknown number of novel classes. We set first 50\% classes as known and remaining 50\% classes as novel.}
    \label{fig:unknown_class}%
    \vspace{-6mm}
\end{figure}

\vspace{-2mm}
\paragraph{\textbf{Unknown Number of Novel Classes:}}
\label{par:unknown}
In our experiments, we assume that the number of novel classes is known in advance. This follows novel class discovery methods \cite{han2019learning,Han2020Automatically,fini2021unified}, and exiting work on open-world SSL \cite{cao2022openworld}. However, this is a limiting assumption since in real-world applications the number of novel classes is rarely known a priori. Therefore, estimating the number of novel classes is crucial for wider adoption. To the best of our knowledge, DTC~\cite{han2019learning} is the only method that proposes a solution for estimating the number of novel classes. However, in our experiments we find that DTC only works for small number of unknown classes and fails to estimate the number of novel classes for CIFAR-100 dataset, where 50\% of classes are novel. Therefore, instead of estimating the number of novel classes by DTC, we analyze the performance of OpenLDN with the assumption that a reasonable method for estimating the number of novel classes is provided. We further assume that the hypothetical method will either overestimate or underestimate the number of novel classes. We conduct two sets of experiments to investigate both these cases. The results are demonstrated in Fig.~\ref{fig:unknown_class}. We observe that performance of OpenLDN is stable over a wide range of estimation errors. We also notice that even with 50\% estimation error OpenLDN outperforms ORCA on CIFAR-100 dataset. In summary, these experiments demonstrate that OpenLDN can be applied in a more realistic setup if a reasonable method for estimating the number of novel classes is available. 

\vspace{-2mm}
\section{Limitations}
\vspace{-2mm}
\label{para:discussion}
In this work, we focus on the more general open-world SSL problem, where the unlabeled set can include unknown class samples. However, our proposed solution is based on a few assumptions. Most notably, following prior works, we assume that the known classes are similar to novel classes, thus they will share some relevant information that can be exploited for discovering novel classes. However, in some extreme cases, this assumption might be violated. Besides, in our entropy regularization term, we encourage the outputs to be uniform. For imbalanced data, this %regularization 
can lead to suboptimal results. One solution is to apply the prior target distribution \cite{rizve2022towards} instead of uniform distribution in the entropy regularization loss. This prompts a new avenue of research that investigate novel methods to estimate prior target distribution in open-world environments.

\vspace{-2mm}
\section{Conclusion}
\vspace{-2mm}
In this work, we propose OpenLDN to solve open-world SSL problem. OpenLDN utilizes a pairwise similarity loss to discover and cluster novel classes by solving a pairwise similarity prediction task. One advantage of our solution is that the proposed pairwise similarity objective does not rely on any additional self-supervised pretraining, but instead, it exploits the information readily available in the labeled set using a bi-level optimization rule. Moreover, our solution brings in a unique perspective towards solving open-world SSL problems by transforming it into a closed-world SSL problem. %once the novel classes are discovered. 
This insight provides the opportunity to leverage all recent advancements in closed-world SSL methods readily in the context of open-world SSL. Finally, we introduce iterative pseudo-labeling as a simple and effective tool to address the noise present in generated pseudo-labels for novel classes without adding any significant computational overhead. OpenLDN is able to outperform the state-of-the-art open-world SSL methods while incurring a lower computational cost. We demonstrate the superior performance of OpenLDN over a wide range of vision datasets.

\clearpage

\appendix
\section*{Appendix}
This appendix includes information about our training algorithm, experimental setup, and further evaluations. We provide our training algorithm in Section \ref{sec:algorithm}. Next, we cover additional implementation details in Section \ref{sec:implementation}. Section \ref{sec:baseline} provides details for the baseline implementations. We conduct experiments within a more restricted environment with limited number of labeled data samples in Section \ref{sec:limited}. We include results for a moderately imbalanced dataset, i.e.\ SVHN, in Section \ref{sec:svhn}. Furthermore, we provide results on three additional datasets (FGVC-Aircraft, Stanford-Cars, and Herbarium19) in Section \ref{sec:three_dataset}. Finally, we discuss the effect of varying the frequency of iterative pseudo-labeling in our proposed OpenLDN approach in Section \ref{sec:ipl}.

\section{OpenLDN Training Algorithm}
\label{sec:algorithm}
We provide OpenLDN training algorithm in Alg.~\ref{alg: alogirthm1}. For OpenLDN training, we require a set of labeled data $\mathbb{S}_L$, and a set of unlabeled data $\mathbb{S}_U$. In addition to this, we need to set the number of maximum iterations for stage-1 (learning to discover novel classes) and stage-2 (closed-world SSL) training: $t_1$ and $t_2$ and also the frequency for iterative pseudo-labeling, $m$. The OpenLDN algorithm outputs trained feature extractor $f_{\Theta}$, and classifier $f_{\Phi}$.

For stage-1 of OpenLDN training, first, we initialize feature extractor, $f_{\Theta}$, classifier, $f_{\Phi}$, and similarity prediction network, $f_{\Omega}$. Next, we sample a batch of labeled examples from known classes, $\mathbf{X}^l$, and their corresponding labels $\mathbf{Y}^l$. We also sample a batch of unlabeled samples from both known and novel classes $\mathbf{X}^u$. After that, to learn to recognise novel classes, we compute $\mathcal{L}_{nov}$ and update the parameters of feature extractor, $f_{\Theta}$, and classifier, $f_{\Phi}$, accordingly. Next, we compute a cross-entropy loss on the labeled examples with these updated parameters and update the parameters of similarity prediction network, $f_{\Omega}$, based on this cross-entropy loss using the proposed bi-level optimization rule. We continue this training procedure for $t_1$ iterations to train the feature extractor $f_{\Theta}$, classifier $f_{\Phi}$, and similarity prediction network, $f_{\Omega}$. We use the trained feature extractor $f_{\Theta}$ and classifier $f_{\Phi}$ for generating pseudo-labels for the subsequent closed-world SSL training.              

For stage-2 of OpenLDN training, first, we generate pseudo-labels, $\mathbb{S}_{PL}$, for novel classes using the trained feature extractor, $f_{\Theta}$, and classifier, $f_{\Phi}$. Next, we select top-k pseudo-labels from each class, $\mathbb{S}_{selected}$. After that, we complement the original set of labeled samples with the selected pseudo-labeled samples, $\mathbb{\tilde{S}}_L$ and also remove the selected pseudo-labeled samples from the unlabeled set to obtain $\mathbb{\tilde{S}}_U$. We re-initialize the feature extractor and classifier for the subsequent closed-world SSL training. Next, we perform closed-world SSL training by sampling a batch of labeled/pseudo-labeled samples $\mathbf{X}^l$, their corresponding labels/pseudo-labels $\mathbf{Y}^l$, and also a batch of unlabeled samples $\mathbf{X}^u$. We update the network parameters based on the appropriate closed-world SSL loss. We repeat the pseudo-label generation process every $m$ iterations to mitigate the impact of noisy pseudo-labels. The stage-2 of OpenLDN ends after $t_2$ iterations and returns the trained feature extractor $f_{\Theta}$, and classifier $f_{\Phi}$.                 
% \vspace{-8mm}
\begin{algorithm}[h]
\caption{OpenLDN training algorithm}
\label{alg: alogirthm1}
\textbf{Input:} Set of labeled data $\mathbb{S}_L$, set of unlabeled data $\mathbb{S}_U$, maximum iterations $t_1$ and $t_2$, and frequency of iterative pseudo-labeling $m$ \\
\textbf{Output:} Trained feature extractor $f_{\Theta}$, and classifier $f_{\Phi}$ \\

\textbf{\hskip 5.7mm Stage-1: Learning to Discover Novel Classes}
\begin{algorithmic}[1]
\State Initialize feature extractor $f_{\Theta}$, classifier $f_{\Phi}$, and similarity prediction network $f_{\Omega}$
\For{$t$ = 1...{$t_1$}} 
  \State $\mathbf{X}^l, \mathbf{Y}^l \gets \mathrm{SampleBatch}(\mathbb{S}_L)$
  \State $\mathbf{X}^u \gets \mathrm{SampleBatch}(\mathbb{S}_U)$ 
  \State $\mathcal{L}\gets\mathcal{L}_{nov}({\Theta}^{(t)}, {\Phi}^{(t)}, {\Omega}^{(t)}, \mathbf{X}^l, \mathbf{X}^u, \mathbf{Y}^l)$ \Comment{Eq. 1}
  \State $({\Theta}^{(t+1)}, {\Phi}^{(t+1)}) \gets ({\Theta}^{(t)}, {\Phi}^{(t)}) - \alpha_{({\Theta}, {\Phi})}\nabla_{({\Theta}, {\Phi})}\mathcal{L}$
  \State $\mathcal{L}\gets\mathcal{L}_{ce}^l({\Theta}^{(t+1)}, {\Phi}^{(t+1)}, \mathbf{X}^l, \mathbf{Y}^l)$
  \State ${\Omega}^{(t+1)} \gets {\Omega}^{(t)} - \alpha_{\Omega}\nabla_{\Omega}\mathcal{L}$
 
\EndFor

\noindent
\textbf{Stage-2: Closed-World SSL Training}
\State $\mathbb{S}_{PL} \gets$ Generate Pseudo-Labels \Comment{Eq. 8} \label{pl1}
\State $\mathbb{S}_{selected} \gets \mathrm{TopK}(\mathbb{S}_{PL})$ \Comment{Select top-k}
\State $\mathbb{\tilde{S}}_L \gets \mathbb{S}_L \cup \mathbb{S}_{selected}$
\State $\mathbb{\tilde{S}}_U \gets \mathbb{S}_U \setminus \mathbb{S}_{selected}$ \label{pl2}
\State Initialize feature extractor $f_{\Theta}$, and classifier $f_{\Phi}$

\For{$t$ = 1...{$t_2$}} 
  \State $\mathbf{X}^l, \mathbf{Y}^l \gets \mathrm{SampleBatch}(\mathbb{\tilde{S}}_L)$
  \State $\mathbf{X}^u \gets \mathrm{SampleBatch}(\mathbb{\tilde{S}}_U)$
\State Update $\Theta$ and $\Phi$ using closed-world SSL loss
  
 \If {$t\%m=0$}
    \State Repeat steps \ref{pl1} to \ref{pl2}
\EndIf
 \EndFor

\State \textbf{return} $f_{\Theta}$, $f_{\Phi}$

\end{algorithmic}
\end{algorithm}

\section{Additional Implementation Details}
\label{sec:implementation}
In this section, we provide additional implementation details of our method. In OpenLDN algorithm, we use standard data augmentations which include random crop, and random horizontal flip for all datasets. To obtain the strongly augmented version of the image, $\mathbf{x}^s$, we use Randaugment \cite{cubuk2020randaugment}. We use the default parameters for Randaugment in all the datasets and set the value of $N$ to 2 and $M$ to 10. Here, $N$ is the number of concurrent random augmentations and $M$ is the magnitude of the selected augmentations in Randaugment. Following the prior works \cite{Han2020Automatically,cao2022openworld}, we also modify the base feature extractor, ResNet-18 \cite{he2016deep}, for CIFAR-10, and CIFAR-100 datasets. To this end, we remove the first max-pooling layer; this helps in dealing with images of smaller resolution (32$\times$32). We also change the first convolutional layer; we set the stride to 1 and the kernel size to 3$\times$3. For Tiny ImageNet dataset experiment, we do not remove the first max-pooling layer. However, we do make the same change to the first convolutional layer as above. We do not modify the network architecture for ImageNet-100, and Oxford-IIIT Pet dataset experiments. To reduce the training time, we downsample the images (train and test) of Oxford-IIIT Pet dataset to 256$\times$256 before applying data augmentation. We use the same downsampling operation for the baseline methods. We do not modify default parameters for Mixmatch \cite{NIPS2019_8749_MixMatch} and UDA \cite{xie2019unsupervised}. Finally, we apply Mixup \cite{zhang2018mixup} augmentation for the labeled and pseudo-labeled data in UDA training. We observe that this change helps the network to generate better pseudo-labels over time. For all the experiments, we report the results from the last epoch.

\section{Baseline Implementation Details}
\label{sec:baseline}
% \vspace{-1mm}
For comparing our results on Tiny ImageNet and Oxford-IIIT Pet datasets, we modify three novel class discovery methods for the open-world SSL problem: DTC \cite{han2019learning}, RankStats \cite{Han2020Automatically}, and UNO\cite{fini2021unified}. The details of these modifications are provided in this section. For DTC\cite{han2019learning}, we extend the unlabeled head to include both known and novel classes (for more details about the unlabeled head please refer to \cite{han2019learning}). Following ORCA \cite{cao2022openworld}, we perform SimCLR pretraining for RankStats\cite{Han2020Automatically}. After that, similar to DTC, we also extend the unlabeled head of RankStats (for more details about the unlabeled head please refer to \cite{Han2020Automatically}). However, neither of these methods in their original formulation assume that the unlabeled data contain samples from known classes. Therefore, extending the unlabeled head to encompass both known and novel classes does not induce any ordering (predefined known class order) for the known classes. Hence, in our evaluation we use Hungarian algorithm \cite{kuhn1955hungarian} to match the known classes from the unlabeled head with the ground-truth labels. Finally, we calculate clustering accuracy on known classes for these two methods. UNO \cite{fini2021unified} is a novel class discovery method which assumes that the unlabeled data only contain samples from novel classes. Therefore, UNO generates pseudo-labels only for novel classes. To extend UNO to open-world SSL setup, we generate pseudo-labels for both known and novel classes. For evaluation, we concatenate the labeled and unlabeled head predictions (for more details about the concatenation strategy please refer to \cite{fini2021unified}). Similar to OpenLDN, we calculate standard accuracy on known classes for UNO. For evaluating novel classes, and the joint task of classifying both known and novel classes we calculate clustering accuracy.

%%%%%%%%%%%%%%%%%%%%%%%%%%%%%%%%%%%%%%%%%%%%%%%%%%%%%%%%%%%%%%%%%%%%%%%%%%%%%%%%%%%%%%%%
\begin{table}[h]
\begin{center}\setlength{\tabcolsep}{2pt}
\small
% \resizebox{0.48\textwidth}{!}{%
\begin{tabular}{lcccc}
\hline

\hline

\hline\\[-3mm]
\textbf{Method} & \textbf{Known} & \textbf{Novel} & \textbf{All}\\
[-3mm]
\\
 \hline

\hline

\hline
FixMatch\cite{sohn2020fixmatch} & $64.3$ & $49.4$ & $47.3$\\
DS$^{3}$L\cite{guo2020safe} & $70.5$ & $46.6$ & $43.5$\\
DTC\cite{han2019learning} & $42.7$ & $31.8$ & $32.4$\\
RankStats\cite{Han2020Automatically} & $71.4$ & $63.9$ & $66.7$\\
UNO\cite{fini2021unified} & $86.5$ & $71.2$ & $78.9$\\
ORCA\cite{cao2022openworld} & $82.8$ & $85.5$ & $84.1$\\
\rowcolor[gray]{.95} {OpenLDN-MixMatch} & $\mathbf{92.4}${\tiny{\textcolor{teal}{$\mathord{\uparrow}9.6$}}} & $\mathbf{93.2}${\tiny{\textcolor{teal}{$\mathord{\uparrow}7.7$}}} & $\mathbf{92.8}${\tiny{\textcolor{teal}{$\mathord{\uparrow}8.7$}}}\\ 
% {\bf OpenLDN-UDA} & $91.5$ & $87.2$ & $89.4$ & $\mathbf{57.8}$ & $33.9$ & $45.8$\\ 
\hline 

\hline

\hline
\end{tabular}
% }
\end{center}
\caption{Average accuracy on \textbf{CIFAR-10}  dataset with 10\% labeled data. We set the first 50\% classes as known and the remaining 50\% classes as novel. The results are averaged over three independent runs.}
\label{tab:cifar10}
% \vspace{-4mm}
% \vspace{-3mm}
\end{table}
%%%%%%%%%%%%%%%%%%%%%%%%%%%%%%%%%%%%%%%%%%%%

% \vspace{-12mm}
\section{Experiments with Limited Number of Labeled Data}
\label{sec:limited}
In this section, we conduct additional experiments on CIFAR-10 and CIFAR-100 datasets with 10\% labeled data. The results on CIFAR-10 dataset are provided in Tab.~\ref{tab:cifar10}. We observe that, similar to the results with 50\% labeled data, OpenLDN significantly outperforms the closed-world SSL method, FixMatch\cite{sohn2020fixmatch}, safe SSL method, DS$^3$L\cite{guo2020safe}, novel class discovery methods, DTC\cite{han2019learning}, RankStats\cite{Han2020Automatically}, and UNO\cite{fini2021unified}. We also observe that OpenLDN outperforms ORCA\cite{cao2022openworld} by 7.7\% on novel classes and 8.7\% on joint task of classifying known and novel classes. These results suggest that the performance gap between OpenLDN and other methods is even higher when working with less amount of annotated data.

%%%%%%%%%%%%%%%%%%%%%%%%%%%%%%%%%%%%%%%%%%%%%%%%%%%%%%%%%%%%%%%%%%%%%%%%%%%%%%%%%%%%%%%%
\begin{table}[h]
\begin{center}\setlength{\tabcolsep}{2pt}

% \resizebox{0.48\textwidth}{!}{%
\begin{tabular}{lccc}
\hline

\hline

\hline\\[-3mm]
\textbf{Method} & \textbf{Known} & \textbf{Novel} & \textbf{All}\\
[-3mm]
\\
 \hline

\hline

\hline
FixMatch\cite{sohn2020fixmatch} & $30.9$ & $18.5$ & $15.3$\\
DS$^{3}$L\cite{guo2020safe} & $33.7$ & $15.8$ & $15.1$\\
DTC\cite{han2019learning} & $22.1$ & $10.5$ & $13.7$\\
RankStats\cite{Han2020Automatically} & $20.4$ & $16.7$ & $17.8$\\
UNO\cite{fini2021unified} & $53.7$ & $33.6$ & $42.7$\\
ORCA\cite{cao2022openworld} & $52.5$ & $31.8$ & $38.6$\\
\rowcolor[gray]{.95} {OpenLDN-MixMatch} & $\mathbf{55.0}${\tiny{\textcolor{teal}{$\mathord{\uparrow}2.5$}}} & $\mathbf{40.0}${\tiny{\textcolor{teal}{$\mathord{\uparrow}8.2$}}} & $\mathbf{47.7}${\tiny{\textcolor{teal}{$\mathord{\uparrow}9.1$}}}\\ 
\hline 

\hline

\hline
\end{tabular}
% }
\end{center}
% \vspace{-4mm}
\small
\caption{Average accuracy on \textbf{CIFAR-100} dataset with 10\% labeled data. We set the first 50\% classes as seen and the remaining 50\% classes as novel. The results are averaged over three independent runs.}
\label{tab:cifar100}
% \vspace{-3mm}
\end{table}
%%%%%%%%%%%%%%%%%%%%%%%%%%%%%%%%%%%%%%%%%%%%

Next, we report the results on CIFAR-100 dataset with 10\% labeled data in Tab.~\ref{tab:cifar100}. Once again we observe performance improvements similar to the ones observed in the CIFAR-10 experiment. On this dataset, OpenLDN outperforms ORCA\cite{cao2022openworld} by 8.2\% on novel classes and 9.1\% on all classes. We draw two conclusions from these results on CIFAR-10 and CIFAR-100 datasts. First, OpenLDN shows larger improvement when the amount of labeled data is limited. This feature is particularly desirable since label efficiency is one of the crucial requirement of a SSL method. Second, we observe that the improvement is higher on CIFAR-100 dataset compared to CIFAR-10 dataset, which suggests that OpenLDN can scale up to challenging datasets (higher number of classes) more efficiently.

%%%%%%%%%%%%%%%%%%%%%%%%%%%%%%%%%%%%%%%%%%%%%%%%%%%%%%%%%%%%%%%%%%%%%%%%%%%%%%%%%%%%%%%
\begin{table}[h]
\begin{center}\setlength{\tabcolsep}{2pt}
\small
% \resizebox{\textwidth}{!}{%
\begin{tabular}{lccc}
\hline

\hline

\hline\\[-3mm]
\textbf{Method}  & \textbf{Known} & \textbf{Novel} & \textbf{All}\\
 [-3mm]
\\
 \hline

\hline

\hline
% DTC\cite{han2019learning} \\
% RankStats\cite{Han2020Automatically} \\
UNO\cite{fini2021unified} & $85.4$ & $74.3$ & $79.0$\\
\rowcolor[gray]{.95} {OpenLDN-MixMatch} & $\mathbf{95.7}${\tiny{\textcolor{teal}{$\mathord{\uparrow}10.3$}}} & $\mathbf{87.2}${\tiny{\textcolor{teal}{$\mathord{\uparrow}12.9$}}} & $\mathbf{92.6}${\tiny{\textcolor{teal}{$\mathord{\uparrow}13.6$}}}\\\hline 

\hline

\hline
\end{tabular}
% }
\end{center}
% \vspace{-4mm}
\caption{Accuracy on \textbf{SVHN}  dataset with 10\% labeled data. We set the first 50\% classes as known and the remaining 50\% classes as novel.}
\label{tab:svhn}
\vspace{-10mm}
\end{table}
%%%%%%%%%%%%%%%%%%%%%%%%%%%%%%%%%%%%%%%%%%%%%%%%%%%%%%%%%%%%

\section{SVHN Experiment}
\label{sec:svhn}

We conduct additional experiment on SVHN\cite{netzer2011reading} dataset. In this experiment, we set the first 5 classes as known and the remaining classes as novel. We consider 10\% data from known classes as labeled. As a baseline, we conduct the same experiment with UNO\cite{fini2021unified}. The results are provided in Tab.~\ref{tab:svhn}. From these results, we observe that OpenLDN outperforms UNO by a large margin. To be specific, OpenLDN improves over UNO by 10.3\% on known classes; an even higher improvement (12.9\%) is observed on novel classes. Finally, on the joint task of classifying both known and novel classes, OpenLDN outperforms UNO by 13.6\%. Results on this dataset provide additional evidence of the effectiveness of OpenLDN which can consistently outperform other existing methods on multiple datasets. It is important to note that the SVHN dataset contains a moderate level of imbalance; the dataset suffers from an imbalance factor\cite{zhou2020bbn} of 2.98. Therefore, the results on this dataset also demonstrates that OpenLDN works reasonably well under moderate imbalance.        

%%%%%%%%%%%%%%%%%%%%%%%%%%%%%%%%%%%%%%%%%%%%%%%%%%%%%%%%%%%%%%%%%%%%%%%%%%%%%%%%%%%%%%%
\begin{table}[h]
\begin{center}\setlength{\tabcolsep}{2pt}
\small
% \resizebox{\textwidth}{!}{%
\begin{tabular}{lccc}
\hline

\hline

\hline\\[-3mm]
\textbf{Method}  & \textbf{FGVC-Aircraf} & \textbf{Stanford-Cars} & \textbf{Herbarium19}\\
 [-3mm]
\\
 \hline

\hline

\hline
% DTC\cite{han2019learning} \\
% RankStats\cite{Han2020Automatically} \\
ORCA\cite{cao2022openworld} & 14.7 & 9.6 & 22.9 \\
\rowcolor[gray]{.95} {OpenLDN-UDA} & $\mathbf{45.7}${\tiny{\textcolor{teal}{$\mathord{\uparrow}31.0$}}} & $\mathbf{38.7}${\tiny{\textcolor{teal}{$\mathord{\uparrow}29.1$}}} & $\mathbf{45.0}${\tiny{\textcolor{teal}{$\mathord{\uparrow}22.1$}}}\\\hline 

\hline

\hline
\end{tabular}
% }
\end{center}
% \vspace{-4mm}
\caption{Accuracy on \textbf{FGVC-Aircraf}, \textbf{Stanford-Cars}, and \textbf{Herbarium19} datasets with 50\% labeled data. We set the first 50\% classes as known and the remaining 50\% classes as novel.}
\label{tab:three_dataset}
\vspace{-10mm}
\end{table}
%%%%%%%%%%%%%%%%%%%%%%%%%%%%%%%%%%%%%%%%%%%%%%%%%%%%%%%%%%%%

\section{Additional Results on FGVC-Aircraft, Stanford-Cars, and Herbarium19 Datasets}
\label{sec:three_dataset}

We conduct additional experiments on FGVC-Aircraft\cite{maji13fine-grained}, Stanford-Cars\cite{KrauseStarkDengFei-Fei_3DRR2013}, and Herbarium19 \cite{tan2019herbarium} dataset. In this experiment, we set the first 50\% classes as known and the remaining classes as novel. We consider 50\% data from known classes as labeled. We compare our results with ORCA. We use ResNet-18 for both ORCA and OpenLDN. The results are provided in Tab.~\ref{tab:three_dataset}. On all three datasets, OpenLDN substantially outperforms ORCA. However, we request readers to interpret the ORCA results with caution since it might be possible to obtain improved results for ORCA with better hyperparameter selection. Overall, these results (\ref{tab:three_dataset}) further validate OpenLDN’s effectiveness on more challenging fine-grained and imbalanced datasets.

\begin{figure*}
\captionsetup[subfloat]{labelformat=empty}
% \vspace{-4mm}
    \centering
    \subfloat[]{{\includegraphics[width=0.32\textwidth]{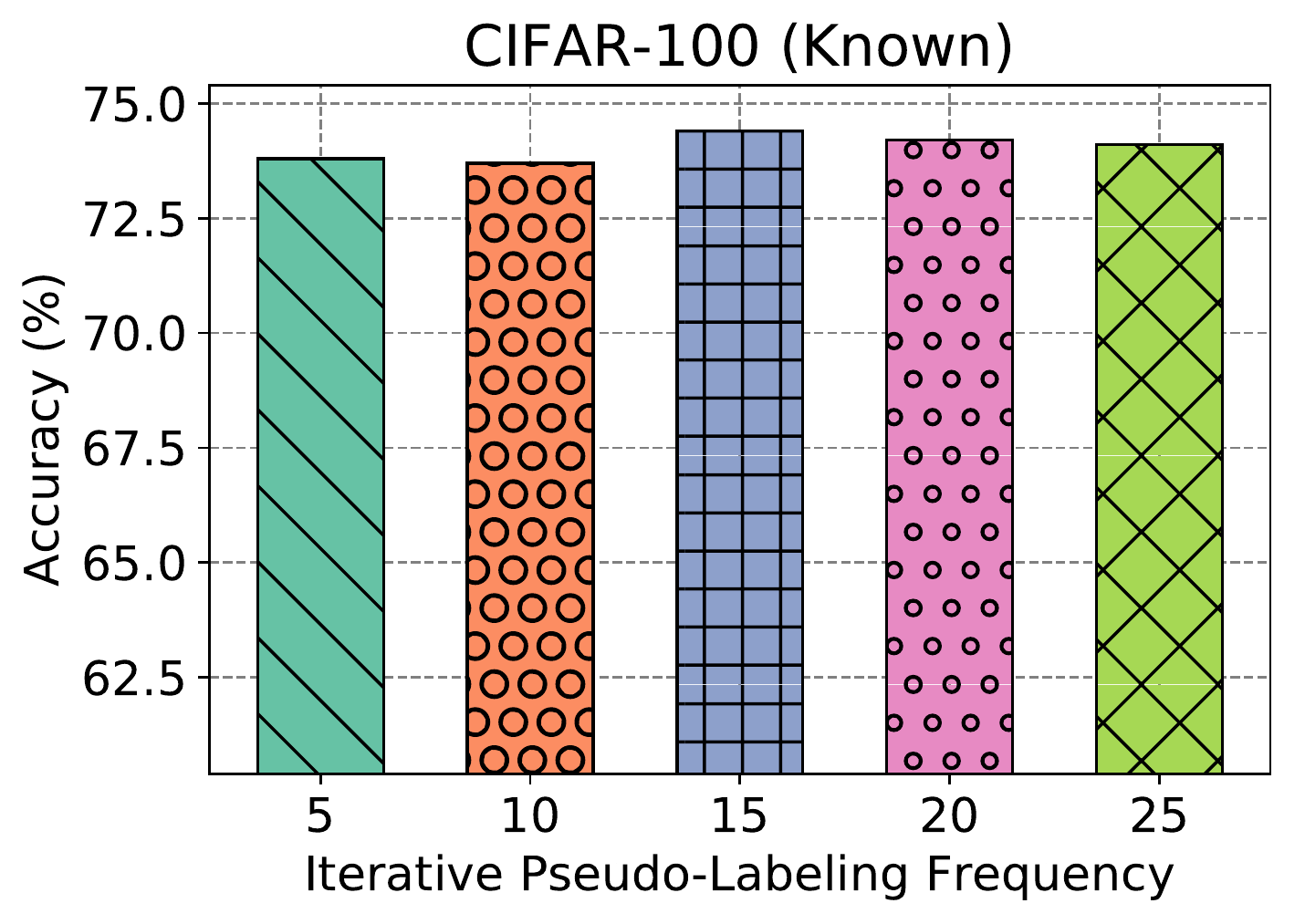}{\label{fig:ipl_known}} }}
    % \qquad
    \subfloat[]{{\includegraphics[width=0.32\textwidth]{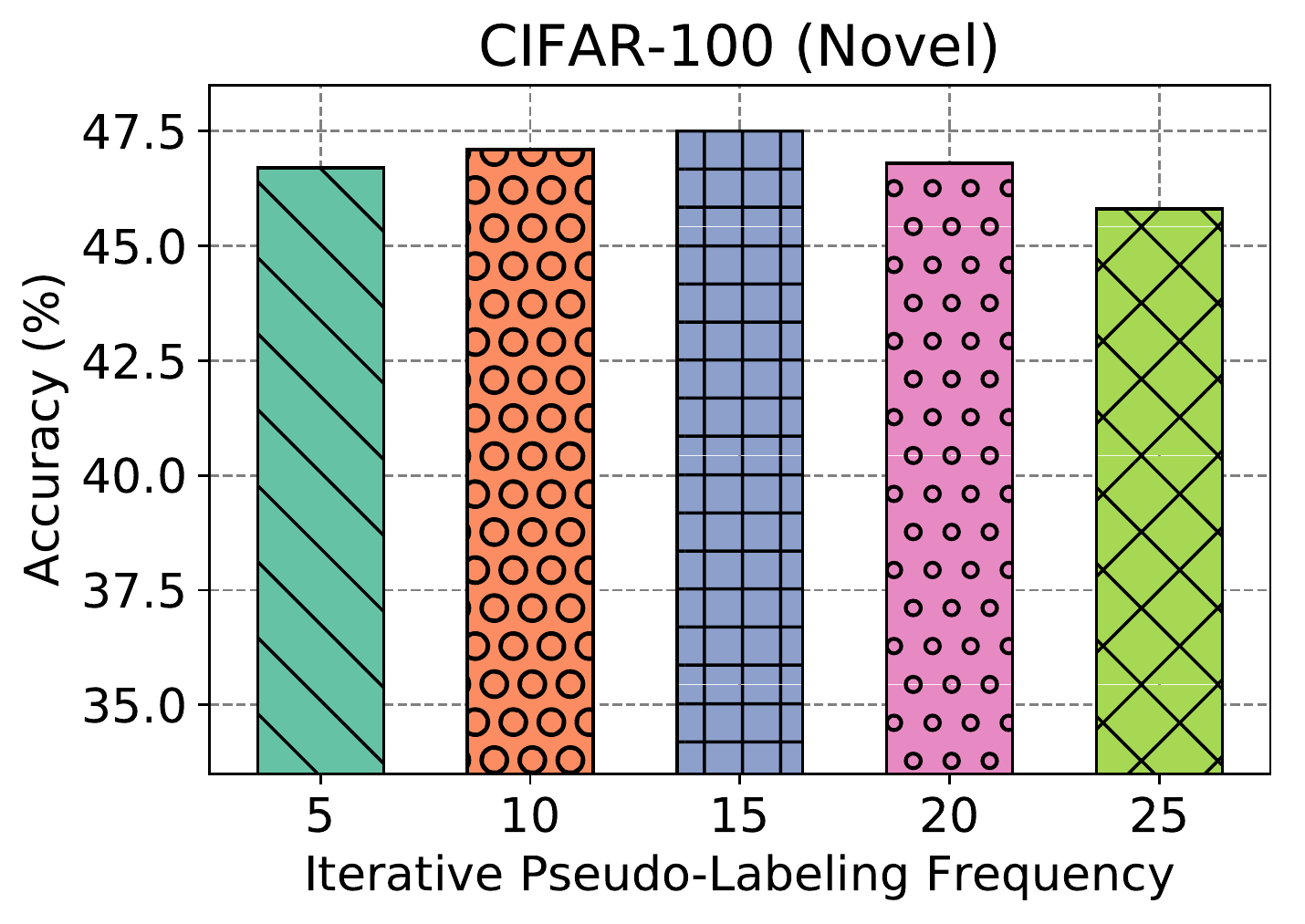}{\label{fig:ipl_novel}} }}
    % \qquad
    \subfloat[]{{\includegraphics[width=0.32\textwidth]{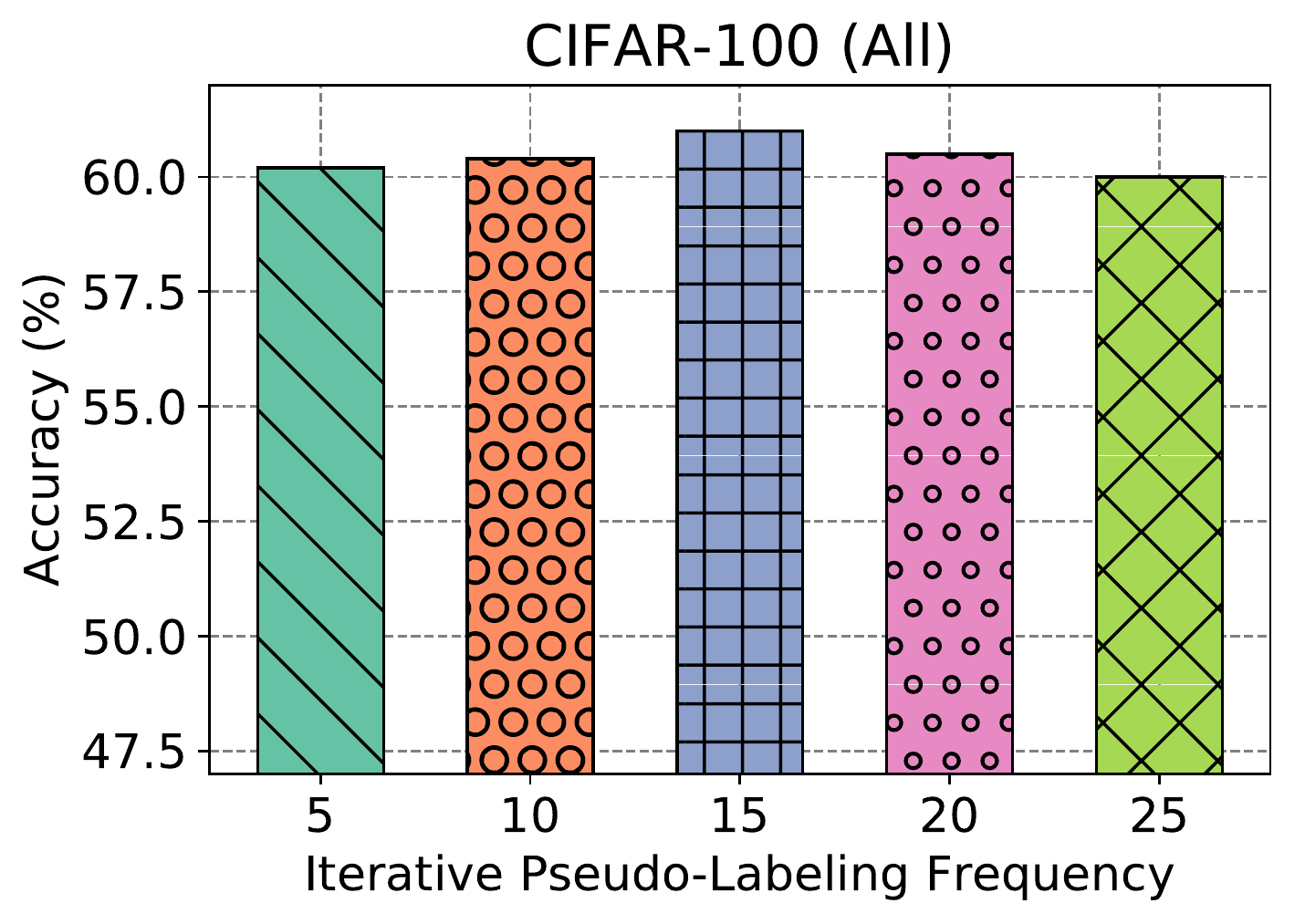}{\label{fig:ipl_all}} }}
\vspace{-4mm}
    \caption{The effect of changing the frequency of iterative pseudo-labeling on final accuracy. These graphs demonstrate classification accuracies on known/novel/all classes for \textbf{CIFAR-100} dataset.} 
    
    \label{fig:ipl}
    % \vspace{-4mm}
\end{figure*}

\section{Frequency of Iterative Pseudo-Labeling}
\label{sec:ipl}

Recall that in OpenLDN algorithm, during the second stage we generate pseudo-labels for novel classes after every 10 epochs. We introduce this iterative pseudo-labeling procedure to mitigate the negative impact of the noise present in the generated pseudo-labels after the novel class discovery phase. Here, we investigate the effect of changing the frequency of iterative pseudo-label generation process. To analyse this effect, we conduct experiments on CIFAR-100 (50\% labeled data). The results are provided in Fig.~\ref{fig:ipl}. We observe that different frequencies for iterative pseudo-labeling lead to similar performance. This suggests that iterative pseudo-labeling is not sensitive to this hyperparameter and can improve the second stage closed-world SSL training irrespective of the frequency used. Besides, Fig.~\ref{fig:ipl} demonstrates that applying a frequency of 15 leads to optimal performance. However, since we do not use any validation set to tune this hyperparameter, in our main experiments, instead, we apply our initial guess and use a frequency of 10.   

\clearpage
\bibliographystyle{eccv2022submission}
\bibliography{eccv2022submission}
\end{document}